\documentclass{article}

\usepackage{microtype}
\usepackage{graphicx}
\usepackage{subfigure}
\usepackage{booktabs} % for professional tables

\usepackage{hyperref}

\usepackage[accepted]{icml2025}

\usepackage{amsmath}
\usepackage{amssymb}
\usepackage{mathtools}
\usepackage{amsthm}
\usepackage[capitalize,noabbrev]{cleveref}
\usepackage{lipsum}
\usepackage{stfloats} 
\theoremstyle{plain}

\theoremstyle{definition}

\theoremstyle{remark}

\usepackage{makecell}

\usepackage[textsize=tiny]{todonotes}

\icmltitlerunning{EMG based speech neuroprosthesis}

\begin{document}

\twocolumn[
\icmltitle{Geometry of orofacial neuromuscular signals: speech articulation decoding using surface electromyography}



\icmlsetsymbol{equal}{*}

\begin{icmlauthorlist}
\icmlauthor{Harshavardhana T. Gowda}{yyy}
\icmlauthor{Zachary D. McNaughton}{zzz}
\icmlauthor{Lee M. Miller}{zzz,ccc,vvv,y}
%\icmlauthor{Firstname3 Lastname3}{comp}
%\icmlauthor{Firstname4 Lastname4}{sch}
%\icmlauthor{Firstname5 Lastname5}{yyy}
%\icmlauthor{Firstname6 Lastname6}{sch,yyy,comp}
%\icmlauthor{Firstname7 Lastname7}{comp}
%\icmlauthor{Firstname8 Lastname8}{sch}
%\icmlauthor{Firstname8 Lastname8}{yyy,comp}
\end{icmlauthorlist}

\icmlaffiliation{yyy}{Department of Electrical and Computer Engineering, University of California, Davis}
\icmlaffiliation{zzz}{Center for Brain and Mind, University of California, Davis}
\icmlaffiliation{ccc}{Department of Neurobiology, Physiology, and Behavior, University of California, Davis}
\icmlaffiliation{vvv}{Department of Otolaryngology/Head and Neck Surgery, University of California, Davis}
\icmlaffiliation{y}{\textcolor{blue}{A version of this article has been published in the Journal of Neural Engineering. \href{https://iopscience.iop.org/article/10.1088/1741-2552/ade7af}{\em J. Neural Eng}}}
%\icmlaffiliation{comp}{Company Name, Location, Country}
%\icmlaffiliation{sch}{School of ZZZ, Institute of WWW, Location, Country}

\icmlcorrespondingauthor{Harshavardhana T. Gowda}{tgharshavardhana@gmail.com}
%\icmlcorrespondingauthor{Firstname2 Lastname2}{first2.last2@www.uk}

\icmlkeywords{Machine Learning, ICML}

\vskip 0.3in
]



\printAffiliationsAndNotice{}  % leave blank if no need to mention equal contribution

\begin{abstract}

\noindent
\textit{Objective.} In this article, we present data and methods for decoding speech articulations using surface electromyogram (EMG) signals. EMG-based speech neuroprostheses offer a promising approach for restoring audible speech in individuals who have lost the ability to speak intelligibly due to laryngectomy, neuromuscular diseases, stroke, or trauma-induced damage (e.g., from radiotherapy) to the speech articulators.

\textit{Approach.} 
To achieve this, we collect EMG signals from the face, jaw, and neck as subjects articulate speech, and we perform EMG-to-speech translation. 

\textit{Main results.} 
Our findings reveal that the manifold of symmetric positive definite (SPD) matrices serves as a natural embedding space for EMG signals. Specifically, we provide an algebraic interpretation of the manifold-valued EMG data using linear transformations, and we analyze and quantify distribution shifts in EMG signals across individuals. 

\textit{Significance.} 
Overall, our approach demonstrates significant potential for developing neural networks that are both data- and parameter-efficient—an important consideration for EMG-based systems, which face challenges in large-scale data collection and operate under limited computational resources on embedded devices.

\end{abstract}

%%%%%%%%%%%%%%%%%%%%%%%%%%%%%%%%%%%%%%%%%%%%%%%%%%%%%%%%%%%%%%%%%%%%%%%%%%%%%%%%%%%%%%%%%%%%%%%%%%%%%%%%%%%%%%%%%%%%%%%%%%%%%%%%%%%%%%%

%%%%%%%%%%%%%%%%%%%%%%%%%%%%%%%%%%%%%%%%%%%%%%%%%%%%%%%%%%%%%%%%%%%%%%%%%%%%%%%%%%%%%%%%%%%%%%%%%%%%%%%%%%%%%%%%%%%%%%%%%%%%%%%%%%%%%
\section{Introduction}
\label{Introduction}

Electromyogram (EMG) signals gathered from the orofacial neuromuscular system during the silent articulation of speech in an alaryngeal manner can be synthesized into personalized audible speech, potentially enabling individuals without vocal function to communicate naturally. EMG based speech neuroprostheses show great promise as such systems can work in many realistic environments, e.g. in noisy backgrounds or with visual occlusion, where traditional methods based on video (such as generating audio from lip movements) may fail. In addition, neural interfaces encode rich information in multiple sensor nodes at different spatial locations and can detect subtle movements and gestures which may not be discernible with video or residual audio signals. 

In this article, we present methods for decoding speech articulation using EMG, focusing on the geometric structure of the data. We demonstrate that EMG signals are naturally embedded on a manifold of SPD matrices. Furthermore, we provide an algebraic interpretation of manifold-valued EMG data and show that EMG signals can be interpreted as a linear transformation of a Euclidean space of dimension $|\mathcal{V}|$, where \( |\mathcal{V}| \) denotes the number of EMG sensor nodes.

We show that all articulations from a given individual result in similar transformations of this space, such that these transformations can be equivalently described using a common approximate eigenbasis matrix. Notably, different individuals have distinct eigenbases. Thus, the distribution shift of EMG signals across individuals can be characterized as a change of basis.

Additionally, we open-source EMG data from 16 subjects performing various orofacial gestures and speech articulations, constituting the largest publicly available speech EMG dataset to date. The data can be downloaded from this open {\href{https://osf.io/ym5jd/}{\textsc{repository}}}, and the code is available on {\small \href{https://github.com/HarshavardhanaTG/geometryOfOrofacialNeuromuscularSystem}{\textsc{GitHub}}}.

%%%%%%%%%%%%%%%%%%%%%%%%%%%%%%%%%%%%%%%%%%%%%%%%%%%%%%%%%%%%%%%%%%%%%%%%%%%%%%%%%%%%%%%%%%%%%%%%%%%%%%%%%%%%%%%%%%%%%%%%%%%%%%%%%%%%%%%%%%

\subsection{Prior work}
\label{Prior work}
A substantial body of prior work (\citeauthor{jou2006towards}, \citeauthor{schultz2010modeling}, \citeauthor{kapur2020non}, \citeauthor{meltzner2018development}, \citeauthor{toth09_interspeech}, \citeauthor{8114359}, and \citeauthor{8578038}) has laid the groundwork for the development of silent speech interfaces. While these studies have been instrumental in shaping the field, they place less emphasis on understanding the {\em data structure} and the implementation of parameter and data-efficient approaches.

The current benchmark in silent speech interfaces is established by \citeauthor{gaddy2020digital, gaddy2021improved}. Using electromyogram (EMG) signals collected during \emph{silently} articulated speech ($E_S$) and \emph{audibly} articulated speech ($E_A$), along with corresponding audio signals ($A$), they develop a recurrent neural transduction model to map time-aligned features of $E_A$ or $E_S$ with $A$.
In their baseline model, joint representations between $E_A$ and $A$ are learned during training, and the model is tested on $E_S$. To improve performance, a refined model aligns $E_S$ with $E_A$ and subsequently uses the aligned features to learn joint representations with $A$.
The methods described above have significant shortcomings that limit their practicality for real-world deployment. They are, \textcircled{\footnotesize 1} the unavailability of good quality $E_A$ and $A$ in individuals who have lost vocal and articulatory functions and \textcircled{\footnotesize 2} the need for a \emph{2x} sized training corpus for learning {\em x} representations (both $E_A$ and $E_S$).

Unlike the works of \citeauthor{gaddy2020digital, gaddy2021improved}, the goal of EMG-based neuroprostheses is to achieve EMG-to-language translation using only \(E_S\), without requiring corresponding \(E_A\) or \(A\), mimicking standard speech-to-text decoding \cite{hori2017joint}. For example, we should be able to decode a silently articulated sentence, \textsc{$<$It's Kind of Fun$>$}, into its corresponding phonemic sequence, \textsc{$<$ih-t-s \textsc{\tiny space} k-ay-n-d \textsc{\tiny space} ah-v \textsc{\tiny space} f-ah-n$>$}, using only \(E_S\). This task is particularly challenging due to the complex multivariate nature of \(E_S\) and the absence of timestamps, as the sentence is articulated silently, making it unclear when a particular word is produced. To accomplish this, we need robust representations of EMG signals. This article describes foundational methods that enable such an EMG-to-language translation. Using the methods described in this article, we demonstrate for the first time that EMG-to-language translation is possible using only \(E_S\), as shown in \citeauthor{gowda2025non}

Finally, none of the previous works have explored whether EMG signals are capable of distinguishing different orofacial movements underlying speech articulation (such as various positions of the tongue and jaw) or whether EMG signals can span the entire English language phonemic space. This article provides data and methods to investigate these fundamental open questions.
In the following sections, we detail methods for obtaining \textit{informative and discriminative representations} of EMG signals and demonstrate that EMG embeddings on the manifold of SPD matrices are meaningful. We show that different orofacial movements, phoneme articulations, and word articulations exhibit structured representations that enable straightforward decoding. Additionally, we analyze and quantify EMG signal distribution shifts across individuals.

%%%%%%%%%%%%%%%%%%%%%%%%%%%%%%%%%%%%%%%%%%%%%%%%%%%%%%%%%%%%%%%%%%%%%%%%%%%%%%%%%%%%%%%%%%%%%%%%%%%%%%%%%%%%%%%%%%%%%%%%%%%%%%%%%%%%%%%%%%%%%%

\section{Methods and materials}
\label{Methods}
A total of 16 subjects (12 female, 4 male) participated in our study. Please refer to the \textit{ethical statement} for details on subject selection criteria. Below, we provide a detailed explanation of the data acquisition protocols and EMG processing methods. The same 12 subjects participated in the experiments described in \textit{sections}~\ref{sec:orofacial}, \ref{sec:phonemes}, and \ref{sec:words}.

\subsection{EMG data acquisition setup}

\begin{figure}[ht] 
    \centering 
    \includegraphics[width=0.3\textwidth]{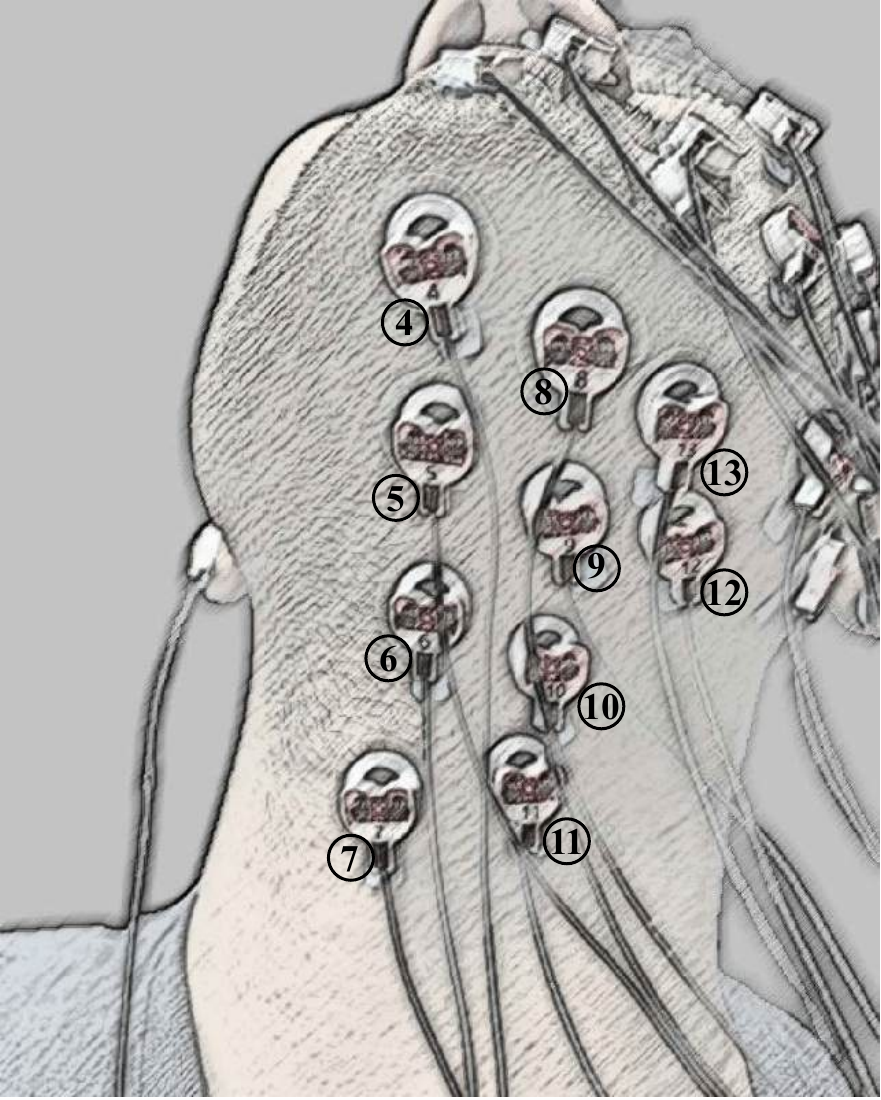} 
    \caption{Placement of electrodes on the neck region.} 
    \label{fig:Face1} 
\end{figure}

\begin{figure}[ht] 
    \centering 
    \includegraphics[width=0.35\textwidth]{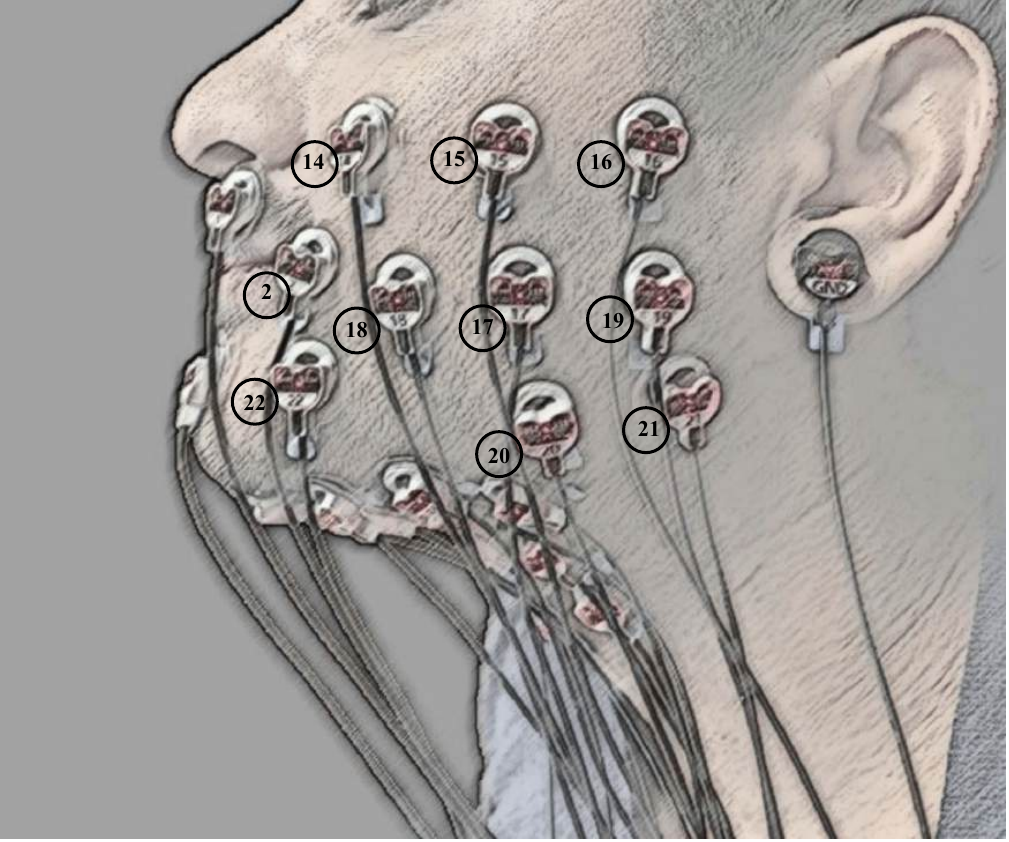} 
    \caption{Placement of electrodes on cheek and lip regions. Electrode {\em 1} is above the upper lip and electrode {\em 3} is below the lower lip.} 
    \label{fig:Face2} 
\end{figure}

We collect EMG signals from twenty-two sites on the neck, chin, jaw, cheek, and lips using monopolar electrodes. An \textsc{actiCHamp Plus} amplifier and associated active electrodes from \textsc{Brain Vision} (\href{https://brainvision.com/products/actichamp-plus/}{Brain Vision}) are used to record EMG signals at 5000 Hertz. To ensure proper contact between the skin surface and electrodes, we use \textsc{SuperVisc}, a high-viscosity electrolyte gel from \textsc{Easycap} (\href{https://shop.easycap.de/products/supervisc}{Easycap}). We develop a software suite in a \textsc{Python} environment to provide visual cues to subjects and to collate and store timestamped data. For time synchronization, we use lab streaming layer (\href{https://labstreaminglayer.org}{LSL}). See figures \ref{fig:Face1} and \ref{fig:Face2} for electrode placement. Besides 22 data electrodes, we also have a \textsc{ground} electrode and a \textsc{reference} electrode. \textsc{ground} electrode is placed on the left ear lobe and the \textsc{reference} electrode is placed on the right ear lobe. Electrode positions and labels are identical for all participants.

Before signal acquisition, participants were briefed on the experimental protocol and seated comfortably in a chair. They were instructed to perform the articulations naturally, as they would in everyday speech. For {\em silent} articulations, participants were asked to speak naturally but inaudibly. Each articulation (or orofacial gesture) was performed within a 1.5-second window, during which the corresponding text was displayed on a graphical user interface (GUI). In the following sections, we provide a detailed description of the various parts of the experiment.
%%%%%%%%%%%%%%%%%%%%%%%%%%%%%%%%%%%%%%%%%%%%%%%%%%%%%%%%%%%%%%%%%%%%%%%%%%%%%%%%%%%%%%%%%%%%%%%%%%%
\subsubsection{EMG of different orofacial movements}
\label{sec:orofacial}

The aim of this experiment is to verify if different orofacial movements underlying speech articulation can be distinguished using EMG. 12 subjects performed 13 different orofacial movements, each repeated 10 times. The orofacial movements are: \textsc{\footnotesize $<$cheeks - puff out$>$}, \textsc{\footnotesize $<$cheeks - suck in$>$}, \textsc{\footnotesize $<$jaw - dropdown$>$}, \textsc{\footnotesize $<$jaw - move backward$>$}, \textsc{\footnotesize $<$jaw - move forward$>$}, \textsc{\footnotesize $<$jaw - move left$>$}, \textsc{\footnotesize $<$jaw - move right$>$}, \textsc{\footnotesize $<$lips - pucker$>$}, \textsc{\footnotesize $<$lips - smile$>$}, \textsc{\footnotesize $<$lips - tuck as if blotting$>$}, \textsc{\footnotesize $<$tongue - back of lower teeth$>$}, \textsc{\footnotesize $<$tongue - back of upper teeth$>$}, \textsc{\footnotesize $<$tongue - the roof of the mouth$>$}. 

These movements are chosen because they encompass a broad range of muscle activations involved in natural speech production, spanning articulation mechanisms such as lip rounding, jaw positioning, and tongue placement, which are crucial for forming different phonemes. These movements correspond to critical articulatory gestures involved in both {\em audible} and {\em silent} speech. Understanding their distinct EMG signatures will help determine whether EMG-based decoding can differentiate between phonemic elements, even in the absence of acoustic output.

\subsubsection{EMG of individual phoneme articulations}
\label{sec:phonemes}
The aim of this experiment is to verify if all English language phonemes can be decoded using EMG when articulated in both {\em audible} and {\em silent} manners. 12 subjects articulated 38 different phonemes, with each phoneme repeated 10 times in both {\em audible} and {\em silent} manners. Phonemes are broadly classified as consonants or vowels. Based on the placement of articulation, consonants are further classified into bilabial, labiodental, dental, alvelor, post vaelor, and approximant consonants. We list all the phonemes below.

{\em Bilabial consonants}: \textsc{\footnotesize $<$baa$>$}, \textsc{\footnotesize $<$paa$>$}, \textsc{\footnotesize $<$maa$>$}; {\em Labiodental consonants}: \textsc{\footnotesize $<$faa$>$}, \textsc{\footnotesize $<$vaa$>$}; {\em Dental consonants}: \textsc{\footnotesize $<$thaa$>$}, \textsc{\footnotesize $<$dhaa$>$}; {\em Alvelor consonants}: \textsc{\footnotesize $<$taa$>$}, \textsc{\footnotesize $<$daa$>$}, \textsc{\footnotesize $<$naa$>$}, \textsc{\footnotesize $<$saa$>$}, \textsc{\footnotesize $<$zaa$>$}; {\em Post vaelor consonants}: \textsc{\footnotesize $<$chaa$>$}, \textsc{\footnotesize $<$shaa$>$}, \textsc{\footnotesize $<$jhaa$>$}, \textsc{\footnotesize $<$zhaa$>$}; {\em Velar consonants}: \textsc{\footnotesize $<$kaa$>$}, \textsc{\footnotesize $<$gaa$>$}, \textsc{\footnotesize $<$ngaa$>$}; {\em Approximant consonants}: \textsc{\footnotesize $<$yaa$>$}, \textsc{\footnotesize $<$raa$>$}, \textsc{\footnotesize $<$laa$>$}, \textsc{\footnotesize $<$waa$>$}; {\em Vowels and Diphthongs}: \textsc{\footnotesize $<$oy$>$} as in bOY, \textsc{\footnotesize $<$ow$>$} as in nOW,  \textsc{\footnotesize $<$ao$>$} as in OUght,  \textsc{\footnotesize $<$aa$>$} as in fAther, \textsc{\footnotesize $<$aw$>$} as in cOW, \textsc{\footnotesize $<$ay$>$} as in mY, \textsc{\footnotesize $<$ae$>$} as in At, \textsc{\footnotesize $<$eh$>$} as in mEt,  \textsc{\footnotesize $<$ey$>$} as in mAte,  \textsc{\footnotesize $<$iy$>$} as in mEET, \textsc{\footnotesize $<$ih$>$} as in It, \textsc{\footnotesize $<$ah$>$} as in HUt, \textsc{\footnotesize $<$uw$>$} as in fOOD, \textsc{\footnotesize $<$er$>$} as in hER, \textsc{\footnotesize $<$uh$>$} as in hOOD. In total, we have 23 consonants and 15 vowels.

Verifying that EMG signals can span the entire English language phonemic space is important because we aim to build a system in which naturally (but inaudibly) articulated speech is decoded using $E_S$ into a corresponding phonemic sequence, similar to standard audio-to-text translation. We demonstrate such a system in \citeauthor{gowda2025non}, and this work lays the foundation for it.

\subsubsection{EMG of individual word articulations}
\label{sec:words}
The aim of this experiment is to verify if we can decode individual words spanning the entire English language phonemic space when words are articulated in both {\em audible} and {\em silent} manners. 12 subjects articulated 36 different words that span the entire English language phonemic space, with each word repeated 10 times in both {\em audible} and {\em silent} manners. The 36 words are: \textsc{\footnotesize $<$eager$>$}, \textsc{\footnotesize $<$lift$>$}, \textsc{\footnotesize $<$eight$>$}, \textsc{\footnotesize $<$edge$>$}, \textsc{\footnotesize $<$cap$>$}, \textsc{\footnotesize $<$matted$>$}, \textsc{\footnotesize $<$tub$>$}, \textsc{\footnotesize $<$box$>$}, \textsc{\footnotesize $<$rune$>$}, \textsc{\footnotesize $<$rook$>$}, \textsc{\footnotesize $<$folder$>$}, \textsc{\footnotesize $<$block$>$}, \textsc{\footnotesize $<$fun$>$}, \textsc{\footnotesize $<$mop$>$}, \textsc{\footnotesize $<$pod$>$}, \textsc{\footnotesize $<$very$>$}, \textsc{\footnotesize $<$went$>$}, \textsc{\footnotesize $<$throat$>$}, \textsc{\footnotesize $<$this$>$}, \textsc{\footnotesize $<$tango$>$}, \textsc{\footnotesize $<$doubt$>$}, \textsc{\footnotesize $<$not$>$}, \textsc{\footnotesize $<$pretty$>$}, \textsc{\footnotesize $<$xerox$>$}, \textsc{\footnotesize $<$rodent$>$}, \textsc{\footnotesize $<$limb$>$}, \textsc{\footnotesize $<$batch$>$}, \textsc{\footnotesize $<$jeep$>$}, \textsc{\footnotesize $<$ship$>$}, \textsc{\footnotesize $<$beige$>$}, \textsc{\footnotesize $<$yes$>$}, \textsc{\footnotesize $<$echo$>$}, \textsc{\footnotesize $<$gold$>$}, \textsc{\footnotesize $<$sing$>$}, \textsc{\footnotesize $<$uh-oh$>$}, \textsc{\footnotesize $<$hiccup$>$}.

Verifying that we can decode words is important because words are composed of a sequence of phonemes (for example, the word \textsc{$<$eager$>$} can be decomposed into the phoneme sequence \textsc{$<$iy-g-er$>$}) and decoding them with high fidelity (e.g., using a recurrent network) demonstrates the potential for decoding continuously articulated sentences.

\subsubsection{Generalizable language-spelling model using NATO phonetic codes}
\label{sec:NATO}

In this experiment, 4 subjects articulate English language sentences using NATO phonetic codes in a {\em silent} manner. For instance, the word \textsc{$<$rainbow$>$} is articulated as \textsc{$<$romeo-alfa-india-november-bravo-oscar-whiskey$>$}. Although this approach does not replicate natural speech, it provides a practical mode of limited communication for individuals who have lost speech articulation capabilities. The experiment is divided into two stages.

In the first stage, subjects silently articulate each of the 26 NATO phonetic codes, repeating each code 20 times. In the second stage, subjects articulate the phonemically balanced \textsc{Rainbow} \cite{fairbanks1960voice} and \textsc{Grandfather} \cite{reilly2012sherlock} passages in a spelled-out manner using NATO phonetic codes. Together, both passages contain 1968 NATO code articulations. Data from the first stage is used for training and validating the decoding models, while data from the second stage is used for testing.

During the second stage, sentences from the passages are displayed word by word, and each character in a word is articulated using the corresponding NATO phonetic code within a 1.5-second articulation window.

\subsection{EMG data preprocessing}

The data collection environment was carefully controlled to eliminate AC electrical interference. EMG signals underwent minimal preprocessing. The signal from the \textsc{reference} channel (the electrode located on the right earlobe) was subtracted from all other EMG data channels. The resulting signals were then bandpass filtered using a third-order Butterworth filter between 80 and 1000~Hz and $z$-normalized along the time dimension for each channel (i.e., each channel is standardized by subtracting its mean and dividing by its standard deviation: $\frac{x - \mu}{\sigma}$, where $x$ denotes the raw signal, $\mu$ is the mean, and $\sigma$ is the standard deviation, all computed along the time dimension). The preprocessed EMG signals were subsequently used to construct a fully connected sensor graph, whose edge matrices are analyzed on the manifold of symmetric positive definite (SPD) matrices. We detail the EMG analysis methods in the following section.

\subsection{EMG data analysis}
EMG signals are collected by a set of sensors $\mathcal{V}$ and are functions of time $t$. A sequence of EMG signals $E_S$ (or $E_A$) corresponding to silently (or audibly) articulated speech, is represented as $E_{S/A} = \textbf{f}_v(t)$ for all $v \in \mathcal{V}$. Here, $\textbf{f}_v(t)$ denotes the EMG signal captured at a sensor node $v$ as a function of time $t$. 

We construct a complete graph $\mathcal{G} = (\mathcal{V}, \mathcal{E}(\tau))$ to represent the functional connectivity of EMG signals, where $\mathcal{E}(\tau)$ denotes the set of edges over a time window $\tau = [t_{\textsc{\small start}}, t_{\textsc{\small end}}]$. The edge weight between two nodes $v_1$ and $v_2 \in \mathcal{V}$ in a time window is defined as $e_{12} = e_{21} = {\textbf{f}}_{v_1}^T {\textbf{f}}_{v_2}$, which corresponds to the covariance of the signals at those nodes during the time interval. Consequently, the edge (adjacency) matrix $\mathcal{E}(\tau)$ is symmetric and positive semi-definite. We convert semi-definite adjacency matrices into definite ones by computing $\mathcal{E} \leftarrow (1 - \eta) \mathcal{E} + \eta \texttt{trace}(\mathcal{E})\mathcal{I}$. Here, $\mathcal{I}$ is an identity matrix whose dimensions are the same as $\mathcal{E}$ and we empirically found that $\eta = 0.1$ suffices for all our data. 
Directly working with SPD matrices using affine-invariant or log-Euclidean metrics \cite{doi:10.1137/050637996} involves computationally expensive operations, such as matrix exponential and matrix logarithm calculations. These operations make mappings between the manifold space and the tangent space, and vice versa, computationally intensive.
To address this, \citeauthor{lin2019riemannian} proposed methods to operate on SPD matrices using Cholesky decomposition. They established a diffeomorphism between the Riemannian manifold of SPD matrices and Cholesky space. In Cholesky space, the computational burden is significantly reduced: logarithmic and exponential computations are restricted to the diagonal elements of the matrix, making them element-wise operations. Additionally, the Fréchet mean (centroid) is derived in a closed form. For an SPD edge matrix $\mathcal{E}$, the corresponding Cholesky decomposition $\mathcal{L} = \textsc{cholesky}(\mathcal{E})$ is such that $\mathcal{E} = \mathcal{L}\mathcal{L}^T$. A matrix $\lfloor\mathcal{L}\rfloor$ is the strictly lower triangular part of the matrix $\mathcal{L}$, and a matrix $\mathbb{D}(\mathcal{L})$ is the diagonal part of the matrix $\mathcal{L}$.
In the following section, we explain in detail, the methods used to analyze SPD matrices. 

\subsubsection{Distance between two SPD matrices $\mathcal{E}_1$ and $\mathcal{E}_2$}
The geodesic distance between two SPD matrices $\mathcal{E}_1$ and $\mathcal{E}_2$ is same as the distance between the corresponding Cholesky matrices $\mathcal{L}_1$ and $\mathcal{L}_2$ and is calculated as 
\begin{align}
    d(\mathcal{L}_1, \mathcal{L}_2) &= \left\{||\lfloor \mathcal{L}_1 \rfloor - \lfloor \mathcal{L}_2\rfloor||_F^2 \right. \nonumber \\
    &\quad + \left. ||\log\mathbb{D}(\mathcal{L}_1) - \log\mathbb{D}(\mathcal{L}_2)||_F^2\right\}^{1/2}, \label{eq:distance}
\end{align}
where $||.||_F$ denotes the Frobenius norm. 
\subsubsection{Fr\'echet mean (centroid) of SPD matrices}

Given a set of ({\em n}) SPD edge matrices $\mathcal{E}$, we first calculate their corresponding Cholesky decompositions $\mathcal{L} = \textsc{cholesky}(\mathcal{E})$, such that $\mathcal{E} = \mathcal{L}\mathcal{L}^T$. Then, the Fr\'echet mean of the Cholesky decomposed matrices $\mathcal{L}$ is given by  

\begin{align}
\mathcal{F}_{\textsc{cholesky}} &= \frac{1}{n} \sum_{i = 1}^{n} \lfloor\mathcal{L}_i\rfloor \hspace{0.3cm} +\nonumber \\
&\quad \exp\left(\frac{1}{n} \sum_{i = 1}^{n} \log(\mathbb{D}(\mathcal{L}_i)\right). \label{eq:mean}
\end{align}

\subsubsection{{\em k}-medoids clustering algorithm}
\label{sec:kMedoids}

Given a set of ({\em n}) SPD edge matrices, $\mathcal{E}$, from different orofacial movements, if the edge matrices $\mathcal{E}$ corresponding to the same movement cluster together while those from different movements remain separate, we can conclude that treating EMG signals as graphs and representing covariance edge matrices $\mathcal{E}$ on the manifold of SPD matrices provides a robust signal representation. This would suggest that such an embedding space is meaningful for naturally distinguishing different orofacial movements. In the {\em results} section, we test whether this is the case for different orofacial movements described in {\em section} \ref{sec:orofacial}.

To identify such natural clusters in an unsupervised manner, we implement the classic {\em k}-medoids algorithm \cite{Kaufman1990PAM} using partitioning around medoids (PAM) heuristic by replacing the Euclidean distance with the geodesic distance defined in equation \ref{eq:distance}.

Whenever we use {\em k}-medoids clustering algorithm, edge matrices $\mathcal{E}(\tau)$ are created using the entire articulation duration of 1.5 seconds.

\subsubsection{Minimum distance to mean algorithm (MDM)}
\label{sec:MDM}

Given a set of ({\em n}) SPD edge matrices, $\mathcal{E}$, from different speech articulations, these matrices may exhibit a structured representation on the manifold of SPD matrices. However, such a structure may not be sufficiently distinct for unsupervised algorithms like {\em k}-medoids clustering to effectively discern. In such cases, supervised algorithms like MDM may be more useful. 

Given {\em m} classification classes and {\em n} training samples, SPD matrices in the training set $\{\mathcal{E}_i^j\}$, where $i\in\{1, 2, ..., $ {\em n}\}and $j\in\{1, 2, ..., $ {\em m}\} are used to construct centroids for each of the {\em m} classes such that the centroid of class $j$ is,
\begin{equation}
    \mathcal{C}^j = \mathbb{E}(\{\textsc{cholesky}({\mathcal{E}}^j)\}), 
\end{equation}
\noindent where the Fr\'echet mean $\mathbb{E}$ is calculated according to equation \ref{eq:mean}. Given a test dataset of SPD matrices $\{\mathcal{T}\}$, $T\in\mathcal{T}$ is assigned to that class whose centroid is nearest to $\textsc{cholesky}(T)$. That is, the class of $T$ is
\begin{equation}
    \arg\min_j d(\textsc{cholesky}(T), \mathcal{C}^j),
\end{equation}
where $d(.)$ is calculated according to equation \ref{eq:distance}.

Whenever we use MDM, edge matrices $\mathcal{E}(\tau)$ are created using the entire articulation duration of 1.5 seconds.

In all cases, when using MDM, we apply the following {\em train-test} data split: out of the 10 total articulation instances of each word or phoneme, 6 are used for training the model, and 4 are used for testing. The articulations in the test set are not present in the training set.

\subsubsection{{\em t-}SNE for EMG data visualization}
\label{sec:tSNE}
We use the {\em t}-SNE ({\em t}-Stochastic Neighbor Embedding) algorithm, adapted from \citeauthor{van2008visualizing}, for data visualization. Unlike standard {\em t}-SNE, which takes vectors as input and uses Euclidean distance, we input edge matrices $\mathcal{E}$ and employ the distance defined in equation \ref{eq:distance}.
%%%%%%%%%%%%%%%%%%%%%%%%%%%%%%%%%%%%%%%%%%%%%%%%%%%%%%%%%%%%%%%%%%%%%%%%%%%%%%%%%%%%%%%%%%%%%%%%%%%%%%%%%%%%%%%%%%%%%%%%%%%%%%%%%%%%%%%%%%%%%
\subsubsection{A neural network for SPD matrix learning}
\label{sec:SPDNet}
We use {\em k}-medoids (defined in {\em section} \ref{sec:kMedoids}) and MDM (defined in {\em section} \ref{sec:MDM}) to test whether EMG signals, when interpreted as a graph arising from the underlying functional connectivity of the neuromuscular system, give rise to edge matrices, $\mathcal{E}$, that are naturally distinguishable on the embedding manifold of SPD matrices (see the {\em results} section). However, to learn more expressive representations of the edge matrices $\mathcal{E}$, we use a neural network as defined in \citeauthor{huang2017riemannian}. This neural network takes SPD matrices $\mathcal{E}$ as input and outputs a classification of the matrices. We describe this neural network in detail in appendix \ref{sec:app1}.
Succinctly, the neural network learns a new matrix, $\mathcal{E}^{(1)}$, from $\mathcal{E}^{(0)}$ via the transformation $\mathcal{E}^{(1)} = W^T\mathcal{E}^{(0)}W$, where $W$ is a full-rank, semi-orthogonal matrix learned by the neural network (when $\mathcal{E}^{(0)}$ is an SPD matrix, $W$ is an orthogonal matrix such that $W^T = W^{-1}$).

Given a set of SPD edge matrices, \(\{\mathcal{E}^{(0)}\}\) (corresponding, for instance, to different speech articulations), if the neural network learns a transformation \( W \) such that the transformed set of matrices, \(\{\mathcal{E}^{(1)}\}\), are all approximately diagonal, then we can conclude that all SPD matrices in \(\{\mathcal{E}^{(0)}\}\) result in a similar linear transformation of the space \( \mathbb{R}^{|\mathcal{V}|} \). That is, this transformation can be equivalently expressed in the corresponding approximate eigenbasis \( W \) such that any \(\mathcal{E}^{(0)}\) can be expressed as  
\[
W\mathcal{E}^{(1)}W^T,
\]
where \(\mathcal{E}^{(1)}\) is approximately diagonal. Therefore, this decomposition can be interpreted as an approximate eigendecomposition, with \( W \) representing the approximate eigenbasis vectors and \(\mathcal{E}^{(1)}\) representing the corresponding approximate eigenvalues.

In the {\em results} section, we test whether this holds true. If so, all articulations from a given individual would share an approximate common eigenbasis. Furthermore, if such an eigenbasis is distinct from those of other individuals, it would suggest that the shift in the EMG signal distribution across individuals arises from a transformation in the underlying {\em basis}.

Whenever we use this method, edge matrices $\mathcal{E}(\tau)$ are created using the entire articulation duration of 1.5 seconds.

For phoneme articulations (as described in {\em section} \ref{sec:phonemes}) and word articulations (as described in {\em section} \ref{sec:words}), when using this method, we apply the following {\em train-test} data split: out of the 10 total articulation instances of each word or phoneme, 6 are used for training the model, and 4 are used for testing. The articulations in the test set are not present in the training set.

For the generalizable language-spelling articulations using NATO phonetic codes (as described in {\em section} \ref{sec:NATO}), we apply the following {\em train-validation-test} data split when using this method.
\begin{itemize}
    \item {\em Train set:} A subset of the articulations from the first stage of the experiment, where each of the 26 NATO phonetic codes was articulated 20 times. We use 16 repetitions of each of the 26 codes, resulting in a total of 416 training instances.
    
    \item {\em Validation set:} A subset of the articulations from the first stage of the experiment, where each of the 26 NATO phonetic codes was articulated 20 times. We use 4 repetitions (not included in the train set) of each of the 26 codes, resulting in a total of 104 validation instances.
    
    \item {\em Test set:} The second stage of the experiment, which consists of 1968 NATO code articulations from both the \textsc{rainbow} and \textsc{grandfather} passages.
\end{itemize}
The model is trained on the train set, and the weights that achieve the best accuracy on the validation set are used for testing on the test set. 

\subsubsection{A recurrent neural network for SPD matrix learning}
\label{sec:recurrent}
The ultimate goal is to translate continuously articulated speech in a {\em silent} manner by afflicted individuals (such as those who have undergone a laryngectomy) into {\em audible} speech using EMG signals ($E_S$). This is very similar to audio-to-text translation, with a {\em key} difference: while audio is a univariate signal that can be modeled as the application of a time-varying filter to a time-varying source signal, EMG signals are multivariate and result from a purely additive combination of muscle activation patterns. Here, we test whether EMG signals contain informative features at a fine-scale resolution of 30 {\em ms} by training a GRU to classify various phonemes (as described in {\em section} \ref{sec:phonemes}), words (as described in {\em section} \ref{sec:words}), and a generalizable language-spelling model using NATO phonetic words (as described in {\em section} \ref{sec:NATO}). This serves as a foundational verification paradigm for the full-fledged EMG-to-language translation framework, which we describe in \citeauthor{gowda2025non}

Standard speech-to-text models \cite{hori2017joint} use recurrent neural networks that process vectorized representations of audio signals. Here, we adapt a recurrent neural network to accept SPD matrices $\mathcal{E}$ as inputs instead of vectors. Specifically, we use the gated recurrent unit (GRU) model described in \citeauthor{10266751} This method modifies the standard Euclidean GRU to accept SPD matrices as input by incorporating techniques from \citeauthor{chen2018neural} and \citeauthor{lou2020neural}

We provide a detailed description of the recurrent model in appendix \ref{sec:app2}.

Whenever we use this method, edge matrices $\mathcal{E}(\tau)$ are generated by slicing the EMG signals using a sliding window with a context size of 150 {\em ms} and a step size of 30 {\em ms}. For an articulation lasting 1.5 seconds, this results in a sequence of 46 edge matrices, which are treated as 46 time steps in the input sequence to the GRU.

The {\em train-validation-test} split is same as described in {\em section} \ref{sec:SPDNet}.

%%%%%%%%%%%%%%%%%%%%%%%%%%%%%%%%%%%%%%%%%%%%%%%%%%%%%%%%%%%%%%%%%%%%%%%%%%%%%%%%%%%%%%%%%%%%%%%%%%%%%%%%%%%%%%%%%%%%%%%%%%%%%%%%
\section{Results}
Here, we present results demonstrating that EMG signals exhibit structured representations on the manifold of SPD matrices. We show that embeddings of different orofacial movements, words, and phonemes on this manifold are meaningful, such that EMG signals corresponding to different articulations naturally form distinct clusters. Furthermore, we apply the methods described in {\em section} \ref{sec:SPDNet} and {\em section} \ref{sec:recurrent} to the generalizable language-spelling corpus using NATO phonetic codes (see {\em section} \ref{sec:NATO}) and demonstrate that high-fidelity decoding of speech articulation is achievable using EMG data.
\subsection{EMG shows structured representation on the manifold of SPD matrices}
When EMG signals are interpreted as graphs defined by the underlying functional connectivity of the orofacial neuromuscular system, the corresponding graph edge matrices exhibit structured representations on the manifold of SPD matrices. Orofacial movements underlying critical articulatory gestures involved in both {\em audible} and {\em silent} speech articulations demonstrate highly structured clustering, which is discernible even using unsupervised algorithms such as {\em k-}medoids clustering ({\em section} \ref{sec:kMedoids}). While complex articulations, such as phonemes and words, are more challenging to distinguish using unsupervised methods, they still exhibit meaningful representations and achieve high decoding accuracy using MDM ({\em section} \ref{sec:MDM}). Decoding accuracy using both {\em k-}medoids and MDM is significantly higher than chance levels, as shown in the following sections.
\subsubsection{Decoding orofacial gestures}
We decode 13 orofacial gestures, as described in {\em section} \ref{sec:orofacial}, using the {\em k-}medoids algorithm ({\em section} \ref{sec:kMedoids}) and present the average decoding accuracy across all 12 subjects in table \ref{tab:unsupervisedOro}. A detailed subject-wise decoding accuracy is provided in appendix \ref{sec:apd3}.
\begin{table}[ht]
\centering
\begin{tabular}{c}
\hline
\begin{tabular}[c]{@{}c@{}}Average decoding accuracy \\of orofacial gestures using \\{\em k-}medoids algorithm\end{tabular} \\ \hline
\textbf{0.736} \\ \hline
\end{tabular}
\caption{Different orofacial gestures are naturally distinguishable on the manifold of SPD matrices. Classification accuracy of 13 orofacial movements described in {\em section} \ref{sec:orofacial}. Chance accuracy is merely $\frac{1}{13} = 0.077$.}
\label{tab:unsupervisedOro}
\end{table}

\subsubsection{Visualizing orofacial gestures using {\em t-}SNE}
\begin{figure}[ht] 
    \centering 
    \includegraphics[width=0.4\textwidth]{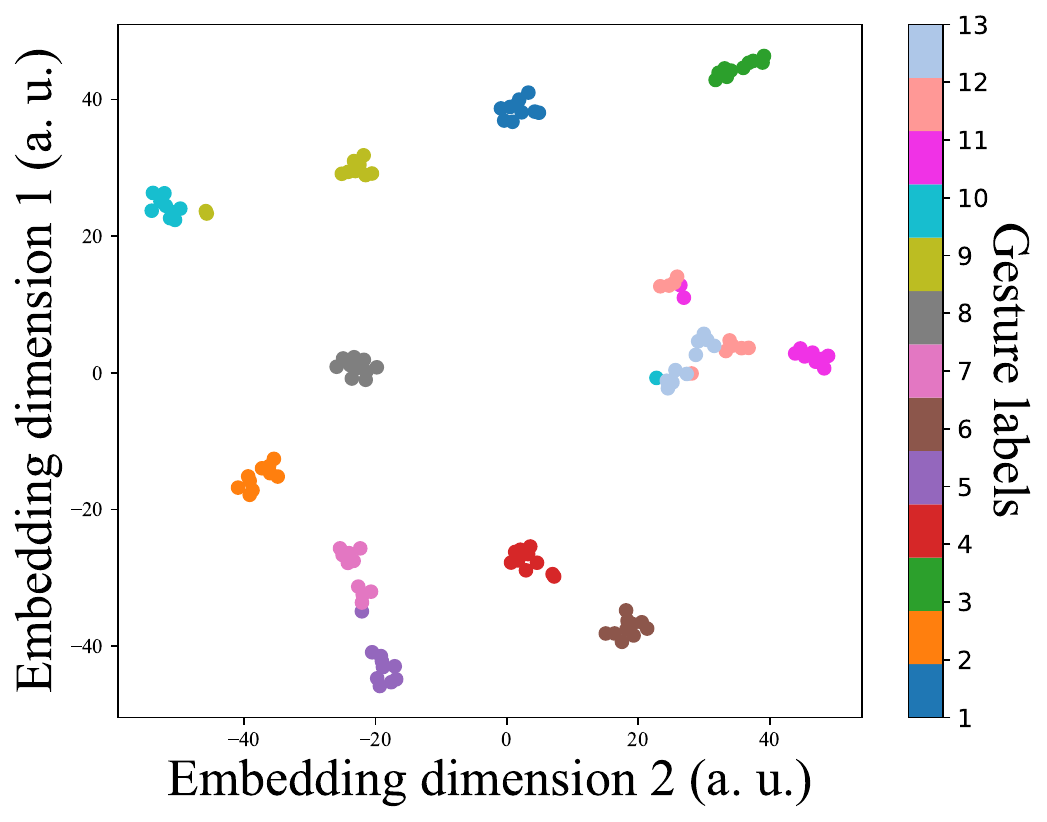} 
    \caption{Different orofacial gestures are naturally distinguishable on the manifold of SPD matrices. {\em t-}SNE of edge matrices of various orofacial movements described in {\em section} \ref{sec:orofacial} for subject {\em 1}. Embedding is colored according to gestures ({\em a.u. - } arbitrary units). } 
    \label{fig:oroIndividual} 
\end{figure}

Here, we visualize SPD edge matrices corresponding to different orofacial gestures using {\em t-}SNE (see {\em section} \ref{sec:tSNE}). As shown in figure \ref{fig:oroIndividual}, different gestures are naturally separated on the manifold.

\subsubsection{Decoding individual phoneme and word articulations}
Individual phoneme articulations (as described in {\em section} \ref{sec:phonemes}) and individual word articulations (as described in {\em section} \ref{sec:words}) exhibit a structured representation that is insufficient for {\em k}-medoids clustering but distinguishable using MDM. This is because speech articulations are more complex compared to orofacial movements. 
We report the average decoding accuracy using MDM across all subjects for both {\em silent} and {\em audible} articulations in tables \ref{tab:phonemesMDM} and \ref{tab:wordsMDM}, respectively. A detailed subject-wise decoding accuracy is provided in appendix \ref{sec:apd4}.

Furthermore, we demonstrate that phonemes and words can be classified with higher accuracy using neural networks described in {\em sections} \ref{sec:SPDNet} and \ref{sec:recurrent}, with results presented in appendices \ref{sec:apd5} and \ref{sec:apd6}, respectively.

{\em Train-test} split for MDM is outlined in {\em section} \ref{sec:MDM}.
\begin{table}[ht]
\centering
\footnotesize
\begin{tabular}{ccc|ccc}
\hline
\multicolumn{3}{c|}{\begin{tabular}[c]{@{}c@{}}Decoding accuracy of \\{\em audible} phonemes\end{tabular}} & \multicolumn{3}{c}{\begin{tabular}[c]{@{}c@{}}Decoding accuracy of \\{\em silent} phonemes\end{tabular}} \\ \hline
 \multicolumn{1}{c|}{\em A} & \multicolumn{1}{c|}{\em C} & \em V & \multicolumn{1}{c|}{\em A} & \multicolumn{1}{c|}{\em C} & \em V \\ \hline

\multicolumn{1}{c|}{\textbf{0.36}} & \multicolumn{1}{c|}{\textbf{0.36}} & \textbf{0.45} & \multicolumn{1}{c|}{\textbf{0.36}} & \multicolumn{1}{c|}{\textbf{0.34}} & \textbf{0.49} \\ \hline
\end{tabular}
\caption{\footnotesize Phoneme articulations are naturally distinguishable on the manifold of SPD matrices. Classification accuracy of phoneme articulations using MDM. It is a 38-way classification problem with chance accuracy of $\frac{1}{38} = 0.026$. We also show classification accuracy for 23 consonant phonemes and 15 vowel phonemes separately. {\em A-}all phonemes, {\em C-}consonants only, {\em V-}vowels only.}
\label{tab:phonemesMDM}
\end{table}

\begin{table}[ht]
\centering
\footnotesize
\begin{tabular}{cl|c}
\hline
\multicolumn{2}{c|}{\begin{tabular}[c]{@{}c@{}}Decoding accuracy of \\ {\em audible} words\end{tabular}} & \multicolumn{1}{c}{\begin{tabular}[c]{@{}c@{}}Decoding accuracy of\\ {\em silent} words\end{tabular}} \\ \hline

\multicolumn{2}{c|}{\textbf {0.544}} & {\textbf {0.439}} \\ \hline
\end{tabular}
\caption{\footnotesize Word articulations are naturally distinguishable on the manifold of SPD matrices. Classification accuracy of word articulations using MDM. It is a 36-way classification problem with chance accuracy of $\frac{1}{36} = 0.028$.}
\label{tab:wordsMDM}
\end{table}

\subsection{Generalizable language-spelling paradigm}
Now, we present the results for the generalizable language-spelling paradigm described in {\em section} \ref{sec:NATO}. Here, we train the neural networks described in {\em sections} \ref{sec:SPDNet} and \ref{sec:recurrent} and report the results. Notably, our {\em training} set consists of only about 10 minutes of data, yet we test on a much larger test set and still achieve decoding accuracies significantly higher than chance level. 

We train the models on 416 NATO phonetic code articulations, validate them on 104 NATO phonetic code articulations, and test them on 1968 articulations from the \textsc{rainbow} and \textsc{grandfather} passages. The top-{\em 5} decoding accuracies are presented in table \ref{tab:NATO}. As observed, these accuracies are significantly higher than the chance-level top-{\em 5} decoding accuracy. The recurrent network described in {\em section} \ref{sec:recurrent} outperforms the network in {\em section} \ref{sec:SPDNet}. This improved performance can be attributed to the recurrent network's ability to model articulations as sequences. A detailed subject-wise decoding accuracy is provided in appendix \ref{sec:apd8}.

\begin{table}[ht]
\centering
\footnotesize
\begin{tabular}{cl|c}
\hline
\multicolumn{2}{c|}{\begin{tabular}[c]{@{}c@{}}Decoding accuracy using \\ model in \\{\em section} \ref{sec:SPDNet}\end{tabular}} & \multicolumn{1}{c}{\begin{tabular}[c]{@{}c@{}}Decoding accuracy using \\ model in \\{\em section} \ref{sec:recurrent}\end{tabular}} \\ \hline

\multicolumn{2}{c|}{\textbf {0.631}} & {\textbf {0.773}} \\ \hline
\end{tabular}
\caption{\footnotesize Top-{\em 5} decoding accuracy of spelled-out \textsc{rainbow} and \textsc{grandfather}  passage articulations using NATO phonetic codes. It is a 26-way classification problem with chance top-{\em 5} accuracy of $0.178$.}
\label{tab:NATO}
\end{table}

\subsection{EMG signal distribution shift across individuals}
\label{sec:dataShift}

EMG signals are significantly affected by distributional shifts across individuals due to inter-subject differences in neural drive and muscle properties \cite{farina2014extraction}, \cite{farina2004extraction}. Factors such as subcutaneous fat thickness, the spatial distribution of muscle fibers, variations in muscle fiber conduction velocity \cite{farina2014extraction}, and contextual elements like electrode placement (which we have ensured is consistent across all subjects to the extent allowed by anatomical and physiological constraints) contribute to this variability. Additionally, neural properties, such as the discharge characteristics of the neural drive, further amplify inter-individual differences in EMG signals \cite{farina2014extraction}. 

Here, we demonstrate that the aforementioned factors lead to covariate signal distribution shifts across individuals, which can be quantified as a change of basis. Specifically, any edge matrix \(\mathcal{E}\) (corresponding to any articulation) from a given individual can be expressed using a common approximate eigenbasis. That is, \(\mathcal{E}\) can be decomposed as \(\mathcal{E} = W\sigma W^T\), where \(\sigma\) is approximately diagonal.  

\begin{figure}[ht] 
    \centering 
    \includegraphics[width=0.4\textwidth]{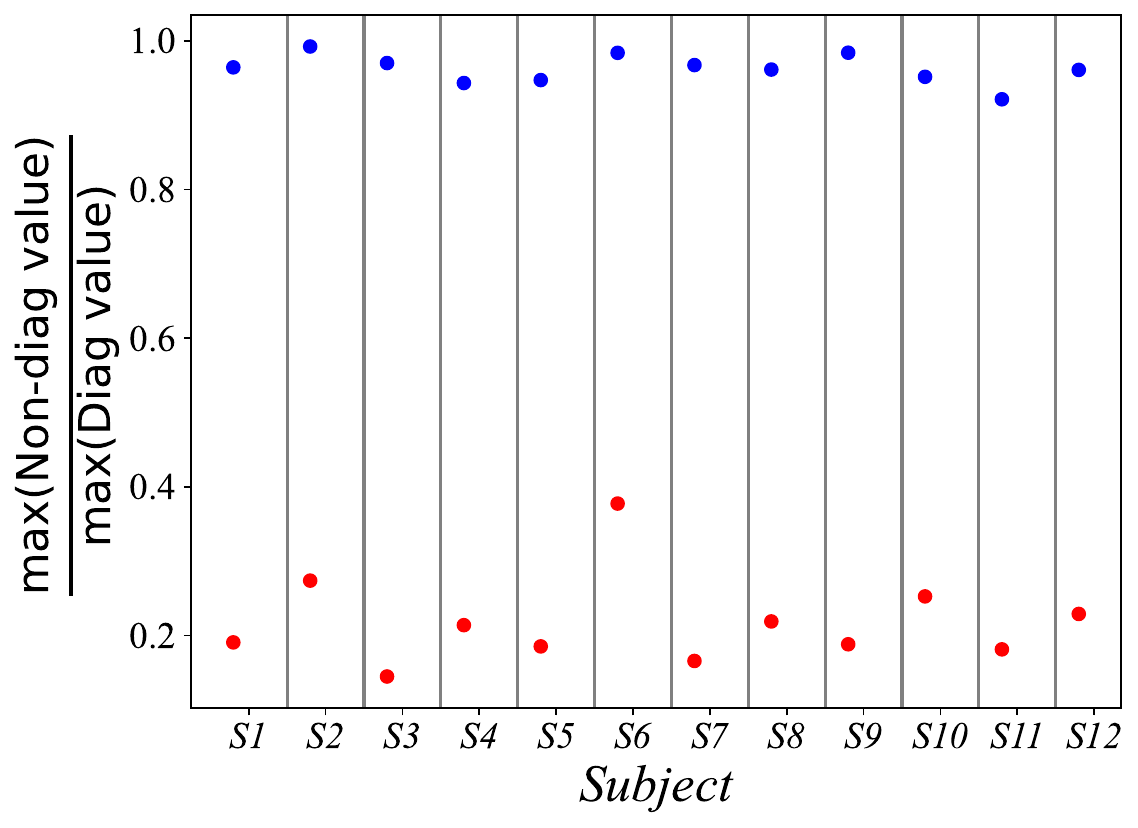} 
    \caption{\footnotesize All edge matrices within an individual can be approximately diagonalized.  \textcolor{blue}{Blue}: average value of $\frac{\max(\textsc{abs}((\textsc{non diag}(\mathcal{E}^{(0)}))}{\max(\textsc{diag}(\mathcal{E}^{(0)}))}$ for all word articulations. \textcolor{red}{Red}: average value of $\frac{\max(\textsc{abs}((\textsc{non diag}(\mathcal{E}^{(1)}))}{\max(\textsc{diag}(\mathcal{E}^{(1)})}$ for all word articulations. As we can see, $\mathcal{E}^{(1)}$ are approximately diagonal compared to $\mathcal{E}^{(0)}$.} 
    \label{fig:nonDiagValues} 
\end{figure}

\begin{figure}[h!] 
    \centering 
    \includegraphics[width=0.4\textwidth]{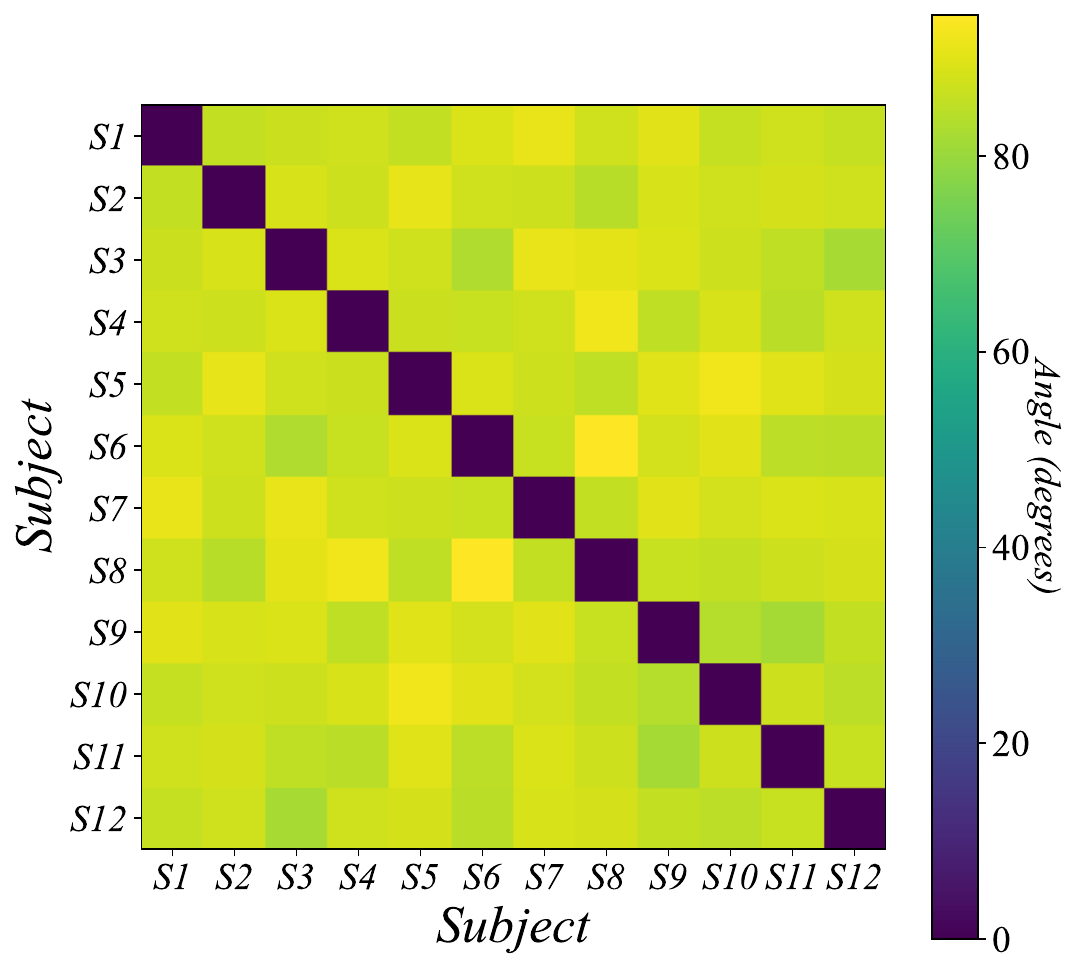} 
    \caption{\footnotesize EMG embeddings of speech articulations on the manifold differ substantially across individuals. Approximate eigenbasis vectors ($Q = W^{(1)}$) are different for different individuals.\\ $\theta = \cos^{-1}\left(\frac{\texttt{trace}(Q_iQ_j^T)}{\sqrt{\texttt{trace}(Q_iQ_i^T)}\sqrt{\texttt{trace}(Q_jQ_j^T)}})\right)$\\ between approximate eigenbasis matrices $Q_i$ and $Q_j$ of different individuals $i$ and $j$.} 
    \label{fig:matrixAngles} 
\end{figure}

We validate this by training an SPD matrix learning neural network separately for each individual (see {\em section} \ref{sec:SPDNet} and appendix \ref{sec:app1}). For instance, we train the neural network to classify different words articulated in an {\em audible} manner ({\em section} \ref{sec:words}: 36 different words spanning the entire phonemic space of the English language, with each word articulated 10 times) and analyze the learned weights that yield the best accuracy on the {\em test} set. The input to the neural network consists of edge matrices corresponding to different word articulations, denoted as \(\mathcal{E}^{(0)}\). In the first layer, the network learns a transformation matrix \(W^{(1)}\) such that 

\[
\mathcal{E}^{(1)} = W^{(1)^T} \mathcal{E}^{(0)} W^{(1)}
\]

is approximately diagonal for all \(\mathcal{E}^{(0)}\). Consequently, any \(\mathcal{E}^{(0)}\) can be decomposed as 

\[
\mathcal{E}^{(0)} = W^{(1)} \mathcal{E}^{(1)} W^{(1)^T},
\]

where \(\sigma = \mathcal{E}^{(1)}\) is approximately diagonal. This demonstrates that all SPD matrices can be represented using an approximate common eigenbasis. However, such an approximate eigenbasis varies across individuals. In figure \ref{fig:nonDiagValues}, we show that matrices $\mathcal{E}^{(1)}$ are more diagonal compared to $\mathcal{E}^{(0)}$ and in figure \ref{fig:matrixAngles}, we show that the transformation matrix $W^{(1)}$ learned by the neural network is different across subjects. 

\subsubsection{Visualizing EMG data distribution shift using {\em t-}SNE}
Here, we visualize SPD edge matrices corresponding to words articulated in an {\em audible} manner ({\em section} \ref{sec:words}) across different subjects using {\em t-}SNE ({\em section} \ref{sec:tSNE}). We observe that, instead of similar words clustering together - for example, the word \textsc{$<$eager$>$} from different subjects does not form a distinct cluster - instead, each subject’s embeddings exhibit a stronger within-subject similarity. Specifically, the embedding of the word \textsc{$<$eager$>$} from subject {\em A} is more similar to other words articulated by subject {\em A} (e.g., \textsc{$<$lift$>$} or \textsc{$<$eight$>$}) than to the same word \textsc{$<$eager$>$} articulated by any other subject.  

We discuss the implications of this observation on the generalization capability of neural networks in the {\em discussion} section. Furthermore, in appendix \ref{sec:apd7}, we demonstrate that this effect persists even for naturally articulated continuous speech when analyzed at a fine phonemic temporal resolution of 100 {\em ms}. This finding has significant implications for the architectural design of EMG-to-language translation models, which we further elucidate in the {\em discussion} section.

\begin{figure}[h!] 
    \centering 
    \includegraphics[width=0.4\textwidth]{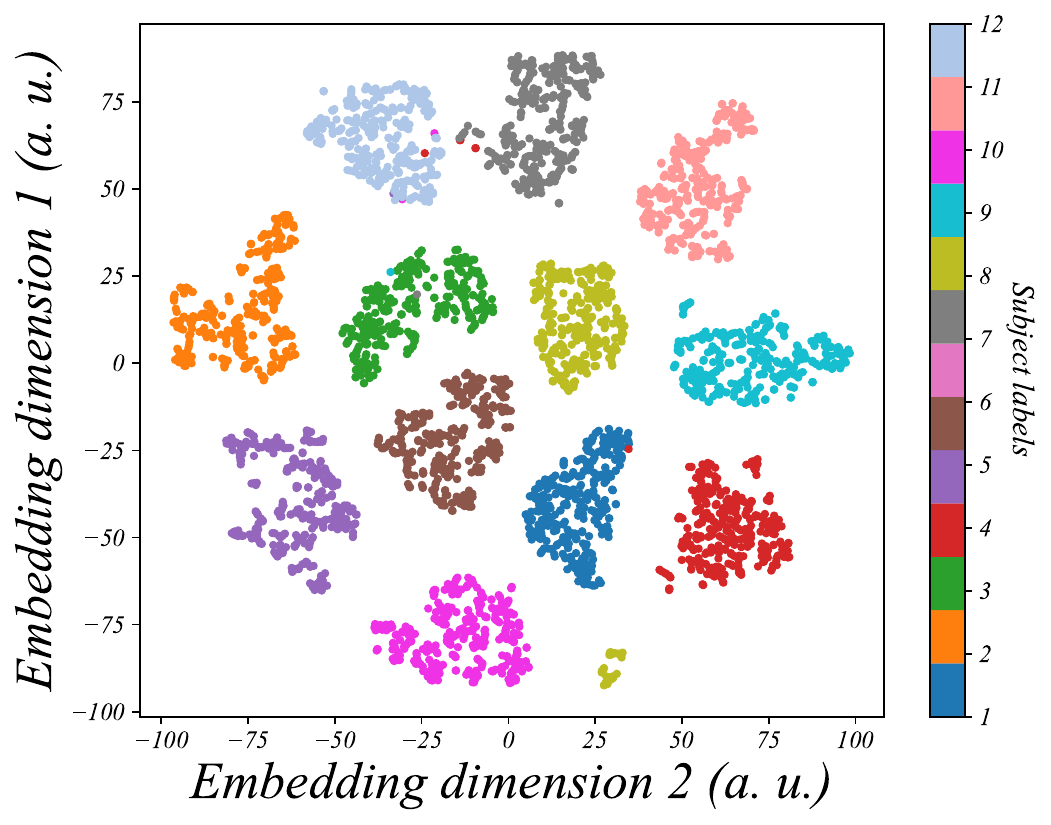} 
    \caption{Articulations from different subjects cluster separately on the manifold of SPD matrices. {\em t-}SNE of edge matrices of word articulations. Embedding is colored according to subjects ({\em a.u. - }arbitrary units).} 
    \label{fig:wordsAllIndividuals} 
\end{figure}

In figure \ref{fig:wordsAllIndividuals}, we show the {\em t-}SNE embedding of {\em audible} word articulations across all 12 subjects. As observed, the embeddings cluster together according to subjects rather than words.

\section{Observations and discussions}
\textcircled{\footnotesize 1} We show that EMG signals, when interpreted as a graph arising from the underlying neuromuscular system, give rise to edge matrices that exhibit a structured representation on the manifold of SPD matrices. We demonstrate this structured representation for orofacial movements underlying speech articulations, as well as for individual word and phoneme articulations (using {\em k-}medoids and MDM algorithms).  

Given that EMG signals exhibit such a structured representation on the manifold, we analyze classification accuracy using more expressive manifold neural networks, including recurrent networks, as described in {\em sections} \ref{sec:SPDNet} and \ref{sec:recurrent}. Using these methods, we show that it is possible to develop limited communication speech neuroprostheses, where sentences are spelled out using NATO codes. These models are easy to train and require only a small amount of data, laying the foundation for a more generalizable EMG-to-language spelling model.

\textcircled{\footnotesize 2} Building on the methods discussed here, we demonstrate—for the first time—that EMG-to-language translation is feasible at the phonemic level with a fine temporal resolution of 20 \textit{ms}, comparable to standard speech-to-text systems, using only EMG signals collected during \textit{silent} speech articulations, as shown in \citeauthor{gowda2025non} On a small-vocabulary dataset containing 67 words, our approach achieves nearly a $2.4\times$ reduction in word error rate (WER) using a model that is $25\times$ smaller. On a large, general-vocabulary dataset, our method yields a 3.5 percentage point improvement in WER, again with a $25\times$ smaller model—all while relying solely on $E_S$, without access to $E_A$ or $A$. Furthermore, our models are computationally efficient, requiring approximately 30 minutes to train on the large-vocabulary dataset and just 2 minutes on the small-vocabulary dataset, in contrast to the 12 hours reported by \citet{gaddy2020digital, gaddy2021improved}.

\textcircled{\footnotesize 3} While EMG-based translation operates on multichannel biosignals reflecting neuromuscular activity, speech processing relies on a fundamentally different signal modality - audio.  Audio signals are univariate, with all articulatory content captured in a single channel. Therefore, they can be viewed as functions sampled on a one-dimensional temporal grid, making deep convolutional neural networks well-suited for learning audio features \cite{baevski2020wav2vec, hsu2021hubert}, as they capture hierarchical representations using varying receptive fields.  

In contrast, EMG signals are multivariate, as they capture muscle activations across multiple sensor nodes. Unlike audio, which is well-represented as a one-dimensional sequence, EMG data is better interpreted as a transformation of the space $\mathbb{R}^{|\mathcal{V}|}$, where $|\mathcal{V}|$ denotes the number of sensors. This distinction arises because EMG captures spatially distributed bioelectrical activity rather than a single waveform. Consequently, instead of treating EMG as a simple sequence, it is more naturally analyzed using structured representations, such as those found on the manifold of SPD matrices.  

We believe this fundamental difference between audio and EMG modalities has important implications for feature extraction and model design, necessitating approaches that go beyond standard convolutional architectures. While direct comparisons for articulatory gestures are lacking - since this work serves as the first such benchmark - previous research \cite{gowda2024topology} has demonstrated that decoding hand gestures using EMG on the SPD manifold outperforms deep learning counterparts defined in the Euclidean domain. Furthermore, by employing the manifold methods described here, we demonstrate that EMG collected from silently articulated speech ($E_S$) can be translated at a phonemic level without relying on corresponding $E_A$ (EMG collected during audible speech) or $A$ (audio) in \citeauthor{gowda2025non} This represents a significant improvement in the training paradigm compared to previous works by \citeauthor{gaddy2020digital, gaddy2021improved} which required both $E_S$ and $E_A$ for model training.

\textcircled{\footnotesize 4}
We analyzed the representations learned by the neural network model in {\em section} \ref{sec:SPDNet} to determine whether the model had learned meaningful representations with respect to the placement and manner of articulation. To this end, we examined phoneme errors to assess whether the model's learned representations correlated well with place and manner of articulation.  

Based on placement and manner of articulation, consonant phonemes are classified into seven groups: bilabial, labiodental, dental, alveolar, post-velar, velar, and approximant consonants. During audible phoneme consonant classification, approximately 44 percent of the errors made by the model in {\em section} \ref{sec:SPDNet} involved misclassifications within the same phoneme group (for example, classifying the bilabial \textsc{$<$paa$>$} as \textsc{$<$baa$>$} or \textsc{$<$maa$>$}, or misclassifying the velar \textsc{$<$zhaa$>$} as \textsc{$<$chaa$>$}, \textsc{$<$shaa$>$}, or \textsc{$<$jhaa$>$}).  

When such misclassifications are considered acceptable, the average consonant phoneme classification accuracy increases from 0.53 ({\em section} \ref{sec:phonemes}) to 0.74, representing a nearly 40 percent improvement. This increase is practically significant, as such systematic errors can be corrected using language models.

\textcircled{\footnotesize 5} Given that the transformation of the space \(\mathbb{R}^{|\mathcal{V}|}\) induced by the edge matrices of EMG from different individuals varies - which can be quantified as a change of basis transformation - and since \(\mathbb{R}^{|\mathcal{V}|}\) can be represented by infinitely many such bases, zero-shot generalization, where a model trained on one group of individuals generalizes to unseen individuals, appears challenging. In future work, we would like to explore if we can design efficient zero-shot or few-shot learning strategies.

\section{Conclusion}
We present an efficient data representation for multivariate orofacial EMG signals and demonstrate that the manifold of SPD matrices provides a meaningful embedding space. Furthermore, we present methods to decode speech articulations using EMG. The open-sourced data and code presented here constitute the largest such dataset to date, establishing a strong foundation for further advancements in the field.  

Several open questions remain to be addressed, including the generalizability of models across individuals - whether such models can be built at all, and if so, what the most efficient strategy for building them would be.

%%%%%%%%%%%%%%%%%%%%%%%%%%%%%%%%%%%%%%%%%%%%%%%%%%%%%%%%%%%%%%%%%%%%%%%%%%%%%%%%%%%%%%%%%%%%%%%%%%%%%%%%%%%%%%%%%%%%%%%%%%%%%%%%%%%%%%%%%%%%%%

\paragraph*{Ethical statement}
Research was conducted in accordance with the principles embodied in the Declaration of Helsinki and in accordance with the University of California, Davis Institutional Review Board Administration protocol 2078695-1. All participants provided written informed consent. Consent was also given for publication of the deidentified data by all participants. Participants were healthy volunteers and were selected from any gender and all ethnic and racial groups. Subjects were aged 18 or above, were able to fully understand spoken and written English, and were capable of following task instructions. Subjects had no skin conditions or wounds where electrodes were placed. Subjects were excluded if they had uncorrected vision problems or neuromotor disorders that prevented them from articulating speech. Children, adults who were unable to consent, and prisoners were not included in the experiments. 

%%%%%%%%%%%%%%%%%%%%%%%%%%%%%%%%%%%%%%%%%%%%%%%%%%%%%%%%%%%%%%%%%%%%%%%%%%%%%%%%%%%%%%%%%%%%%%%%%%%%%%%%%%%%%%%%%%

\paragraph*{Acknowledgments}
This work was supported by awards to Lee M. Miller from: Accenture, through the Accenture Labs Digital Experiences group; CITRIS and the Banatao Institute at the University of California; the University of California Davis School of Medicine (Cultivating Team Science Award); the University of California Davis Academic Senate; a UC Davis Science Translation and Innovative Research (STAIR) Grant; and the Child Family Fund for the Center for Mind and Brain. \\\\
Harshavardhana T. Gowda is supported by Neuralstorm Fellowship, NSF NRT Award No. 2152260 and Ellis Fund
administered by the University of California, Davis.\\

\paragraph*{Conflict of interest}
H. T. Gowda and L. M. Miller are inventors on intellectual property related to silent speech owned by the Regents of University of California, not presently licensed.

\paragraph*{Author contributions}
\begin{itemize}
    \item Harshavardhana T. Gowda: Mathematical formulation, concepts development, data analysis, experiment design,
data collection software design, data collection, manuscript preparation.
\item Zachary D. McNaughton: Data collection.
\item Lee M. Miller: Concepts development and \\manuscript preparation.
\end{itemize}

%\bibliography{example_paper}
%\bibliographystyle{icml2025}

%%%%%%%%%%%%%%%%%%%%%%%%%%%%%%%%%%%%%%%%%%%%%%%%%%%%%%%%%%%%%%%%%%%%%%%%%%%%%%%
%%%%%%%%%%%%%%%%%%%%%%%%%%%%%%%%%%%%%%%%%%%%%%%%%%%%%%%%%%%%%%%%%%%%%%%%%%%%%%%
% APPENDIX
%%%%%%%%%%%%%%%%%%%%%%%%%%%%%%%%%%%%%%%%%%%%%%%%%%%%%%%%%%%%%%%%%%%%%%%%%%%%%%%
%%%%%%%%%%%%%%%%%%%%%%%%%%%%%%%%%%%%%%%%%%%%%%%%%%%%%%%%%%%%%%%%%%%%%%%%%%%%%%%
\appendix
\section{Neural networks for learning SPD matrices}
Here, we describe in detail the neural networks presented in {\em sections} \ref{sec:SPDNet} and \ref{sec:recurrent}.

\subsection{A neural network on Stiefel manifold}
\label{sec:app1}
\begin{figure}[ht] 
    \centering 
    \includegraphics[width=0.4\textwidth]{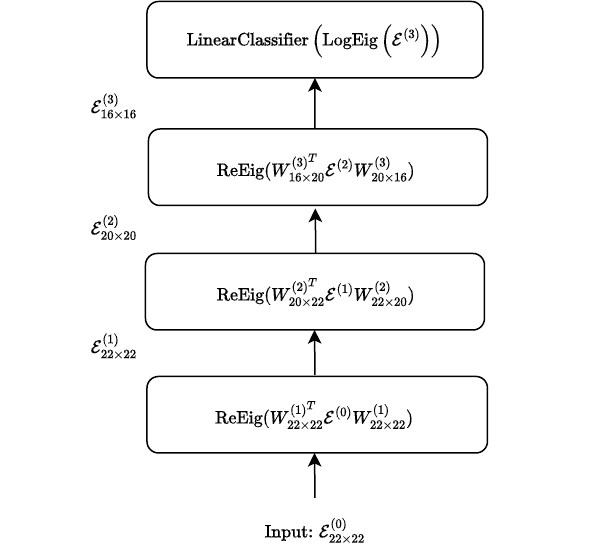} 
    \caption{Neural network architecture for SPD matrix learning.} 
    \label{fig:CNN} 
\end{figure}

The neural network based on the methods described by \citeauthor{huang2017riemannian} is shown in figure \ref{fig:CNN}. A brief description of this network was provided in {\em section} \ref{sec:SPDNet}.

The neural network architecture for learning discriminative SPD matrix representations is made of three layer types. First, a layer is defined by $\mathcal{E}_k = W_k^T\mathcal{E}_{k - 1}W_k$, where an SPD edge matrix $\mathcal{E}_{k-1}$ of dimensions $c_{k-1}\times c_{k-1}$ is input to $k$-th layer. $W_k$ is of dimensions $c_{k-1}\times c_{k}$ giving rise to $\mathcal{E}_k$ of dimensions $c_k\times c_k$. $W_k$ is constrained to be a full-rank semi-orthogonal matrix on a compact Stiefel manifold such that $W_k^TW_k = I$. Second, a non-linear layer is defined by $\mathcal{E}_k$ = $U_{k-1}\max(\epsilon I, \Sigma_{k - 1})U_{k - 1}^T$ (\texttt{ReEig} layer). Third, a layer to map SPD matrices from the manifold space to its tangent space (so that the Euclidean operations can be applied) is defined by $\mathcal{E}_k = U_{k-1}\log(\Sigma_{k - 1})U_{k - 1}^T$ (\texttt{LogEig} layer). $U_{k-1}$, $\Sigma_{k-1}$ are obtained by eigendecomposition of matrix $\mathcal{E}_{k-1}$ and $\epsilon$ is a small constant $>0$.\\

For backpropagation, the gradient of the loss function $L$ with respect to $W_k$, when restricted to the tangent space of the Stiefel manifold (denoted by $\mathbb{R}^{c_{k-1}\times c_k}$, the space of all full-rank matrices of dimension $c_{k-1}\times c_k$) is given by 
\[
\nabla L^{(k)}_{W_k} = \nabla L^{(k)}_{W_k(\texttt{Euclidean})} - W_k\nabla L^{T^{(k)}}_{W_k(\texttt{Euclidean})}W_k.
\]
The gradient is updated as 
\[
W_k\leftarrow W_k - \lambda\nabla L^{(k)}_{W_k},
\]
where $\lambda$ is the learning rate. $W_k$ is then mapped back to the Stiefel manifold (from the tangent space) via orthogonalization. We use Gram-Schmidt method for matrix orthogonalization. Refer to \citeauthor{huang2017riemannian} for backpropagation formulae through non-linear and tangent space mapping layers.
%%%%%%%%%%%%%%%%%%%%%%%%%%%%%%%%%%%%%%%%%%%
\subsection{Manifold recurrent neural network}
\begin{figure}[ht] 
    \centering 
    \includegraphics[width=0.49\textwidth]{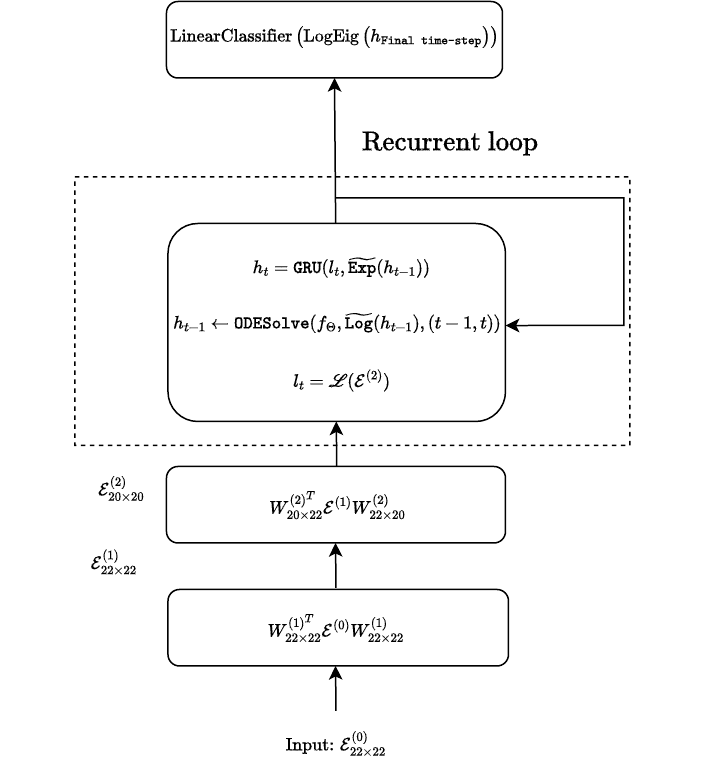} 
    \caption{Recurrent neural network architecture for SPD matrix learning.} 
    \label{fig:cnnRnn} 
\end{figure}

The recurrent neural network based on the methods described by \citeauthor{10266751} is integrated with the methods given by \citeauthor{huang2017riemannian} and is shown in figure \ref{fig:cnnRnn}. A brief description of this network was provided in {\em section} \ref{sec:recurrent}.

\label{sec:app2}

Given a set of SPD edge matrices $\mathcal{E}(\tau)$ over different time windows $\tau$, we first calculate their corresponding Cholesky decompositions $\mathcal{L}(\tau) = \textsc{cholesky}(\mathcal{E}(\tau))$, such that $\mathcal{E}(\tau) = \mathcal{L}(\tau)\mathcal{L(\tau)}^T$. 

Gates of GRU as defined by \citeauthor{10266751} are given below. 

Update-gate $z_{\tau}$ at time-step $\tau$ is
\begin{multline}
    z_\tau = \textsc{sigmoid}(w_z\lfloor \mathcal{L}_\tau\rfloor + u_z\lfloor h_{\tau-1}\rfloor + b_z) +\\\textsc{sigmoid}(b_{z'}[\exp(w_{z'}\log(\mathbb{D}(\mathcal{L}_\tau)) + \\ u_{z'}\log(\mathbb{D}(h_{\tau-1}))]),
    \label{eq:updateGate}
\end{multline}
where $w_z$, $u_z$, $b_z$, $w_{z'}$, and $u_{z'}$ are real weights and $b_{z'}$ is a real positive weight.

Reset-gate $r_\tau$ at time-step $\tau$ is 
\begin{multline}
  r_\tau = \textsc{sigmoid}(w_r\lfloor \mathcal{L}_\tau\rfloor + u_r\lfloor h_{\tau-1}\rfloor + b_r) +\\\textsc{sigmoid}(b_{r'}[\exp(w_{r'}\log(\mathbb{D}(\mathcal{L}_\tau)) + \\u_{r'}\log(\mathbb{D}(h_{\tau-1}))]),
  \label{eq:resetGate}
\end{multline}
where $w_r$, $u_r$, $b_r$, $w_{r'}$, and $u_{r'}$ are real weights and $b_{r'}$ is a real positive weight.

Candidate-activation vector $\hat{h}_\tau$ is 
\begin{multline}
    \hat{h}_\tau=\textsc{tanh}(w_h\lfloor \mathcal{L}_\tau\rfloor + u_h(\lfloor r_\tau\rfloor * \lfloor h_{\tau - 1}\rfloor) + b_h)+\\
    \textsc{softplus}(b_{h'}\exp(w_{h'}\log(\mathbb{D}(\mathcal{L}_\tau))\\+u_{h'}\log(\mathbb{D}(r_\tau)*\mathbb{D}(h_{\tau-1})))),
    \label{eq:candidateActivation}
\end{multline}
where $w_h$, $u_h$, $b_h$, $w_{h'}$, and $u_{h'}$ are real weights and $b_{h'}$ is a real positive weight.

Output vector $h_\tau$ is
\begin{multline}
    h_\tau = (1-\lfloor z_\tau\rfloor) * \lfloor h_{\tau-1}\rfloor + \lfloor z_\tau\rfloor * \lfloor \hat{h_\tau}\rfloor +\\
    \exp((1 - \mathbb{D}(z_\tau))*\log(\mathbb{D}(h_{\tau-1}))+\\\mathbb{D}(z_\tau)*\log(\mathbb{D}(\hat{h}_\tau))).
    \label{eq:outputVector}
\end{multline}
In the above equations, $h_{\tau-1}$ is the hidden-state at time-step $\tau-1$.

We define an implicit layer solved using neural ordinary differential equations. The dynamics $f$ of EMG data is modeled by a neural network with parameters $\Theta$. The output state $h_\tau$ is updated as,
\begin{multline}
        h_{\tau - 1}\leftarrow \textsc{ODESolve}(f_{\Theta}, \widetilde{\textsc{log}}(h_{\tau-1}), (\tau-1, \tau))\\
        h_\tau = \textsc{GRU}(\mathcal{L}_\tau, \widetilde{\textsc{exp}}(h_{\tau-1})),
\end{multline}
where $\widetilde{\textsc{log}}$ is the logarithm mapping from the manifold space of SPD matrices to its tangent space and $\widetilde{\textsc{exp}}$ is its inverse operation as defined by \citeauthor{lin2019riemannian}. $\textsc{GRU}$ is a gated recurrent unit whose gates are given by equations \ref{eq:updateGate} - \ref{eq:outputVector}.

%%%%%%%%%%%%%%%%%%%%%%%%%%%%%%%%%%%%%%%%%%
\section{Detailed results}
Here, we present detailed subject-wise results.

\subsection{Orofacial gesture decoding}
In table \ref{tab:oroSubjectWise}, we present the subject-wise decoding accuracy for 13 different orofacial gestures (described in {\em section} \ref{sec:orofacial}) using the {\em k}-medoids algorithm, as described in {\em section} \ref{sec:kMedoids}. As observed, the edge matrices corresponding to different gestures exhibit a structured representation on the manifold of SPD matrices across subjects.

\label{sec:apd3}
\begin{table}[ht]
\footnotesize
\centering
\begin{tabular}{|c|c|}
\hline
Subject number & \begin{tabular}[c]{@{}c@{}}Decoding accuracy
of \\orofacial gestures using\\
{\em k}-medoids algorithm\end{tabular} \\ \hline
1             & 0.877          \\ \hline
2             & 0.862          \\ \hline
3             & 0.677          \\ \hline
4             & 0.638          \\ \hline
5             & 0.654          \\ \hline
6             & 0.915          \\ \hline
7             & 0.554          \\ \hline
8             & 0.515          \\ \hline
9             & 0.846          \\ \hline
10            & 0.854          \\ \hline
11            & 0.731          \\ \hline
12            & 0.715          \\ \hline
\textbf{Mean} & \textbf{0.736} \\ \hline
\end{tabular}
\caption{Different orofacial gestures are naturally distinguishable on the manifold of SPD matrices. Classification accuracy of 13 orofacial movements described in {\em section} \ref{sec:orofacial}. Chance accuracy is merely $\frac{1}{13} = 0.077$.}
\label{tab:oroSubjectWise}
\end{table}

\subsection{Phoneme and word articulation decoding}
Here, we present the decoding accuracy of phoneme and word articulations using methods described in {\em sections} \ref{sec:MDM}, \ref{sec:SPDNet}, and \ref{sec:recurrent}.
\subsubsection{Phonemes and words decoding using MDM}
\label{sec:apd4}

In table \ref{tab:phonemesMDMSubjectWise}, we present the decoding accuracy of phoneme articulations (described in {\em section} \ref{sec:phonemes}), and in table \ref{tab:wordsMDMSubjectWise}, we present the decoding accuracy of word articulations (described in {\em section} \ref{sec:words}), using MDM (described in {\em section} \ref{sec:MDM}).
As observed, the edge matrices corresponding to different word and phoneme articulations exhibit a structured representation on the manifold of SPD matrices across subjects. Phoneme articulations from subject 11 were corrupted and are unavailable.

\begin{table}[ht]
\centering
\footnotesize
\begin{tabular}{|c|ccc|ccc|}
\hline
\begin{tabular}[c]{@{}c@{}}Subject \\ number\end{tabular} & \multicolumn{3}{c|}{\begin{tabular}[c]{@{}c@{}}Decoding accuracy \\of {\em audible} phonemes\end{tabular}} & \multicolumn{3}{c|}{\begin{tabular}[c]{@{}c@{}}Decoding accuracy \\of {\em silent} phonemes\end{tabular}} \\ \hline
 & \multicolumn{1}{c|}{\em A} & \multicolumn{1}{c|}{\em C} & \em V & \multicolumn{1}{c|}{\em A} & \multicolumn{1}{c|}{\em C} & \em V \\ \hline
1 & \multicolumn{1}{c|}{0.38} & \multicolumn{1}{c|}{0.34} & 0.52 & \multicolumn{1}{c|}{0.5} & \multicolumn{1}{c|}{0.47} & 0.57 \\ \hline
2 & \multicolumn{1}{c|}{0.51} & \multicolumn{1}{c|}{0.47} & 0.67 & \multicolumn{1}{c|}{0.49} & \multicolumn{1}{c|}{0.42} & 0.63 \\ \hline
3 & \multicolumn{1}{c|}{0.26} & \multicolumn{1}{c|}{0.18} & 0.45 & \multicolumn{1}{c|}{0.28} & \multicolumn{1}{c|}{0.29} & 0.3 \\ \hline
4 & \multicolumn{1}{c|}{0.20} & \multicolumn{1}{c|}{0.23} & 0.25 & \multicolumn{1}{c|}{0.33} & \multicolumn{1}{c|}{0.38} & 0.43 \\ \hline
5 & \multicolumn{1}{c|}{0.47} & \multicolumn{1}{c|}{0.52} & 0.53 & \multicolumn{1}{c|}{0.45} & \multicolumn{1}{c|}{0.41} & 0.58 \\ \hline
6 & \multicolumn{1}{c|}{0.5} & \multicolumn{1}{c|}{0.46} & 0.58 & \multicolumn{1}{c|}{0.45} & \multicolumn{1}{c|}{0.41} & 0.58 \\ \hline
7 & \multicolumn{1}{c|}{0.19} & \multicolumn{1}{c|}{0.25} & 0.2 & \multicolumn{1}{c|}{0.27} & \multicolumn{1}{c|}{0.28} & 0.33 \\ \hline
8 & \multicolumn{1}{c|}{0.33} & \multicolumn{1}{c|}{0.46} & 0.33 & \multicolumn{1}{c|}{0.30} & \multicolumn{1}{c|}{0.33} & 0.42 \\ \hline
9 & \multicolumn{1}{c|}{0.36} & \multicolumn{1}{c|}{0.26} & 0.6 & \multicolumn{1}{c|}{0.23} & \multicolumn{1}{c|}{0.17} & 0.5 \\ \hline
10 & \multicolumn{1}{c|}{0.41} & \multicolumn{1}{c|}{0.41} & 0.48 & \multicolumn{1}{c|}{0.31} & \multicolumn{1}{c|}{0.23} & 0.63 \\ \hline
12 & \multicolumn{1}{c|}{0.35} & \multicolumn{1}{c|}{0.41} & 0.38 & \multicolumn{1}{c|}{0.32} & \multicolumn{1}{c|}{0.39} & 0.4 \\ \hline
\textbf{Mean} & \multicolumn{1}{c|}{\textbf{0.36}} & \multicolumn{1}{c|}{\textbf{0.36}} & \textbf{0.45} & \multicolumn{1}{c|}{\textbf{0.36}} & \multicolumn{1}{c|}{\textbf{0.34}} & \textbf{0.49} \\ \hline
\end{tabular}
\caption{\footnotesize Phoneme articulations are naturally distinguishable on the manifold of SPD matrices. Classification accuracy of phoneme articulations using MDM. It is a 38-way classification problem with chance accuracy of $\frac{1}{38} = 0.026$. We also show classification accuracy for 23 consonant phonemes and 15 vowel phonemes separately. {\em A-}all phonemes, {\em C-}consonants only, {\em V-}vowels only.}
\label{tab:phonemesMDMSubjectWise}
\end{table}

\begin{table}[ht]
\centering
\footnotesize
\begin{tabular}{|c|cl|c|}
\hline
\begin{tabular}[c]{@{}c@{}}Subject \\ number\end{tabular} & \multicolumn{2}{c|}{\begin{tabular}[c]{@{}c@{}}Decoding accuracy \\of {\em audible} words\end{tabular}} & \multicolumn{1}{l|}{\begin{tabular}[c]{@{}l@{}}Decoding accuracy \\of {\em silent} words\end{tabular}} \\ \hline
1 & \multicolumn{2}{c|}{0.632} & 0.389 \\ \hline
2 & \multicolumn{2}{c|}{0.660} & 0.590 \\ \hline
3 & \multicolumn{2}{c|}{0.389} & 0.264 \\ \hline
4 & \multicolumn{2}{c|}{0.264} & 0.243 \\ \hline
5 & \multicolumn{2}{c|}{0.632} & 0.708 \\ \hline
6 & \multicolumn{2}{c|}{0.688} & 0.569 \\ \hline
7 & \multicolumn{2}{c|}{0.382} & 0.125 \\ \hline
8 & \multicolumn{2}{c|}{0.549} & 0.472 \\ \hline
9 & \multicolumn{2}{c|}{0.611} & 0.382 \\ \hline
10 & \multicolumn{2}{c|}{0.625} & 0.444 \\ \hline
11 & \multicolumn{2}{c|}{0.587} & 0.674 \\ \hline
12 & \multicolumn{2}{c|}{0.5} & 0.403 \\ \hline
\textbf{Mean} & \multicolumn{2}{c|}{\textbf {0.544}} & {\textbf {0.439}} \\ \hline
\end{tabular}
\caption{\footnotesize Word articulations are naturally distinguishable on the manifold of SPD matrices. Classification accuracy of word articulations using MDM. It is a 36-way classification problem with chance accuracy of $\frac{1}{36} = 0.028$.}
\label{tab:wordsMDMSubjectWise}
\end{table}
%%%%%%%%%%%%%%%%%%%%%%%%%%%%%%%%%%%%%%%%%%%%%%%%%%%%%%%%%%%%%%%%%%%%%%%%%%%%%%%%%%%%%%%%%%%%%%%%%%%%%%%%%%%%%%%%%%%%%%%
\onecolumn
\subsubsection{Phonemes and words decoding using method in {\em section} \ref{sec:SPDNet}}
\label{sec:apd5}
While we use MDM (described in {\em section} \ref{sec:MDM}) to demonstrate that phoneme and word articulations exhibit a structured representation on the manifold of SPD matrices, we can leverage the neural network described in {\em section} \ref{sec:SPDNet} to learn more expressive and distinguishable features. As shown here, decoding accuracy using the neural network is higher than that of MDM for both word and phoneme articulations. Table \ref{tab:phonemeCNN} presents the decoding accuracy of phoneme articulations, while table \ref{tab:wordsCNN} reports the decoding accuracy of word articulations. Phoneme articulations from subject 11 were corrupted and are unavailable.
\begin{table*}[h]
\centering
\footnotesize
\begin{tabular}{|c|cccccc|}
\hline
\begin{tabular}[c]{@{}c@{}}Subject number\end{tabular} & \multicolumn{6}{c|}{Decoding accuracy of phoneme articulations using model  in {\em section} \ref{sec:SPDNet}} \\ \hline
 & \multicolumn{3}{c|}{{\em Audible} phonemes} & \multicolumn{3}{c|}{{\em Silent} phonemes} \\ \hline
 & \multicolumn{1}{c|}{\textit{A}} & \multicolumn{1}{c|}{\textit{C}} & \multicolumn{1}{c|}{\textit{V}} & \multicolumn{1}{c|}{\textit{A}} & \multicolumn{1}{c|}{\textit{C}} & \textit{V} \\ \hline
1 & \multicolumn{1}{c|}{0.574±0.012} & \multicolumn{1}{c|}{0.632±0.015} & \multicolumn{1}{c|}{0.725±0.013} & \multicolumn{1}{c|}{0.574±0.012} & \multicolumn{1}{c|}{0.647±0.016} & 0.678±0.013 \\ \hline
2 & \multicolumn{1}{c|}{0.679±0.012} & \multicolumn{1}{c|}{0.645±0.012} & \multicolumn{1}{c|}{0.857±0.008} & \multicolumn{1}{c|}{0.679±0.012} & \multicolumn{1}{c|}{0.575±0.011} & 0.798±0.016 \\ \hline
3 & \multicolumn{1}{c|}{0.404±0.008} & \multicolumn{1}{c|}{0.428±0.013} & \multicolumn{1}{c|}{0.632±0.014} & \multicolumn{1}{c|}{0.404±0.008} & \multicolumn{1}{c|}{0.393±0.009} & 0.412±0.008 \\ \hline
4 & \multicolumn{1}{c|}{0.310±0.014} & \multicolumn{1}{c|}{0.370±0.015} & \multicolumn{1}{c|}{0.487±0.023} & \multicolumn{1}{c|}{0.310±0.014} & \multicolumn{1}{c|}{0.467±0.014} & 0.670±0.015 \\ \hline
5 & \multicolumn{1}{c|}{0.637±0.007} & \multicolumn{1}{c|}{0.715±0.008} & \multicolumn{1}{c|}{0.660±0.008} & \multicolumn{1}{c|}{0.637±0.007} & \multicolumn{1}{c|}{0.566±0.013} & 0.667±0.017 \\ \hline
6 & \multicolumn{1}{c|}{0.577±0.012} & \multicolumn{1}{c|}{0.542±0.011} & \multicolumn{1}{c|}{0.748±0.019} & \multicolumn{1}{c|}{0.577±0.012} & \multicolumn{1}{c|}{0.550±0.012} & 0.688±0.013 \\ \hline
7 & \multicolumn{1}{c|}{0.313±0.007} & \multicolumn{1}{c|}{0.386±0.007} & \multicolumn{1}{c|}{0.373±0.011} & \multicolumn{1}{c|}{0.313±0.007} & \multicolumn{1}{c|}{0.334±0.014} & 0.408±0.013 \\ \hline
8 & \multicolumn{1}{c|}{0.348±0.036} & \multicolumn{1}{c|}{0.518±0.021} & \multicolumn{1}{c|}{0.563±0.026} & \multicolumn{1}{c|}{0.348±0.036} & \multicolumn{1}{c|}{0.390±0.010} & 0.605±0.021 \\ \hline
9 & \multicolumn{1}{c|}{0.522±0.009} & \multicolumn{1}{c|}{0.479±0.014} & \multicolumn{1}{c|}{0.738±0.017} & \multicolumn{1}{c|}{0.522±0.009} & \multicolumn{1}{c|}{0.305±0.009} & 0.723±0.013 \\ \hline
10 & \multicolumn{1}{c|}{0.603±0.015} & \multicolumn{1}{c|}{0.620±0.014} & \multicolumn{1}{c|}{0.762±0.027} & \multicolumn{1}{c|}{0.603±0.015} & \multicolumn{1}{c|}{0.390±0.017} & 0.755±0.011 \\ \hline
12 & \multicolumn{1}{c|}{0.485±0.011} & \multicolumn{1}{c|}{0.550±0.012} & \multicolumn{1}{c|}{0.577±0.019} & \multicolumn{1}{c|}{0.485±0.011} & \multicolumn{1}{c|}{0.455±0.012} & 0.597±0.015 \\ \hline
 \textbf{Mean}& \multicolumn{1}{c|}{\textbf{0.496}} & \multicolumn{1}{c|}{\textbf{0.535}} & \multicolumn{1}{c|}{\textbf{0.647}} & \multicolumn{1}{c|}{\textbf{0.447}} & \multicolumn{1}{c|}{\textbf{0.461}} & \textbf{0.636} \\ \hline
\end{tabular}
\caption{Classification accuracy of phoneme articulations using model in {\em section} \ref{sec:SPDNet}.  We also show classification accuracy for 23 consonant phonemes and 15 vowel phonemes separately. {\em A-}all phonemes, {\em C-}consonants only, {\em V-}vowels only. It is a 38-way classification problem with chance accuracy of $\frac{1}{38} = 0.026$. Accuracies are averaged over 10 random seeds.}
\label{tab:phonemeCNN}
\end{table*}

\begin{table*}[ht]
\centering
\footnotesize
\begin{tabular}{|c|p{3cm}|p{3cm}|}
\hline
\begin{tabular}[c]{@{}c@{}}Subject number\end{tabular} 
& \multicolumn{2}{c|}{\makecell{Decoding accuracy of word \\articulations using model in \\{\em section} \ref{sec:SPDNet}}} \\ \hline
 & \multicolumn{1}{c|}{{\em Audible} words} & \multicolumn{1}{c|}{{\em Silent} words}  \\ \hline
1 & \multicolumn{1}{c|}{0.747±0.010} & \multicolumn{1}{c|}{0.529±0.010}  \\ \hline
2 & \multicolumn{1}{c|}{0.883±0.009} & \multicolumn{1}{c|}{0.765±0.008}  \\ \hline
3 & \multicolumn{1}{c|}{0.599±0.010} & \multicolumn{1}{c|}{0.443±0.008}  \\ \hline
4 & \multicolumn{1}{c|}{0.340±0.011} & \multicolumn{1}{c|}{0.423±0.010}  \\ \hline
5 & \multicolumn{1}{c|}{0.821±0.012} & \multicolumn{1}{c|}{0.792±0.012} \\ \hline
6 & \multicolumn{1}{c|}{0.807±0.011} & \multicolumn{1}{c|}{0.692±0.014}  \\ \hline
7 & \multicolumn{1}{c|}{0.421±0.010} & \multicolumn{1}{c|}{0.260±0.006}  \\ \hline
8 & \multicolumn{1}{c|}{0.694±0.012} & \multicolumn{1}{c|}{0.578±0.012}  \\ \hline
9 & \multicolumn{1}{c|}{0.722±0.007} & \multicolumn{1}{c|}{0.458±0.010}  \\ \hline
10 & \multicolumn{1}{c|}{0.803±0.010} & \multicolumn{1}{c|}{0.722±0.011}  \\ \hline
11 & \multicolumn{1}{c|}{0.708±0.008} & \multicolumn{1}{c|}{0.772±0.006}  \\ \hline
12 & \multicolumn{1}{c|}{0.634±0.012} & \multicolumn{1}{c|}{0.570±0.014}  \\ \hline
\textbf{Mean} & \multicolumn{1}{c|}{\textbf{0.681}} & \multicolumn{1}{c|}{\textbf{0.584}}  \\ \hline
\end{tabular}
\caption{Classification accuracy of word articulations using model in section \ref{sec:SPDNet}. It is a 36-way classification problem with chance accuracy of $\frac{1}{36} = 0.028$. Accuracies are averaged over 10 random seeds.}
\label{tab:wordsCNN}
\end{table*}

%%%%%%%%%%%%%%%%%%%%%%%%%%%%%%%%%%%%%%%%%%%%%%%%%%%%%%%%%%%%%%%%%%%%%%%%%%%%%%%%%%%%%%%%%%%%%%%%%%%%%%%%%%%%%%%%%%%%%%
\newpage
\subsubsection{Phonemes and words decoding using method in {\em section} \ref{sec:recurrent}}
\label{sec:apd6}
While we use MDM (described in {\em section} \ref{sec:MDM}) to demonstrate that phoneme and word articulations exhibit a structured representation on the manifold of SPD matrices, we can leverage the neural network described in {\em section} \ref{sec:recurrent} to learn more expressive and distinguishable features. As shown here, decoding accuracy using the neural network is higher than that of MDM for both word and phoneme articulations. Table \ref{tab:phonemeRNN} presents the decoding accuracy of phoneme articulations, while table \ref{tab:wordsRNN} reports the decoding accuracy of word articulations. Phoneme articulations from subject 11 were corrupted and are unavailable.

\begin{table*}[ht]
\centering
\footnotesize
\begin{tabular}{|c|cccccc|}
\hline
\begin{tabular}[c]{@{}c@{}}Subject \\ number\end{tabular} & \multicolumn{6}{c|}{Decoding accuracy of word articulations using model in {\em section} \ref{sec:recurrent}} \\ \hline
 & \multicolumn{3}{c|}{{\em Audible} phonemes} & \multicolumn{3}{c|}{{\em Silent} phonemes} \\ \hline
 & \multicolumn{1}{c|}{\textit{A}} & \multicolumn{1}{c|}{\textit{C}} & \multicolumn{1}{c|}{\textit{V}} & \multicolumn{1}{c|}{\textit{A}} & \multicolumn{1}{c|}{\textit{C}} & \textit{V} \\ \hline
1 & \multicolumn{1}{c|}{0.651±0.030} & \multicolumn{1}{c|}{0.699±0.033} & \multicolumn{1}{c|}{0.637±0.027} & \multicolumn{1}{c|}{0.574±0.014} & \multicolumn{1}{c|}{0.596±0.030} & 0.662±0.044 \\ \hline
2 & \multicolumn{1}{c|}{0.659±0.023} & \multicolumn{1}{c|}{0.625±0.026} & \multicolumn{1}{c|}{0.815±0.046} & \multicolumn{1}{c|}{0.590±0.017} & \multicolumn{1}{c|}{0.516±0.020} & 0.780±0.031 \\ \hline
3 & \multicolumn{1}{c|}{0.398±0.030} & \multicolumn{1}{c|}{0.370±0.049} & \multicolumn{1}{c|}{0.498±0.038} & \multicolumn{1}{c|}{0.328±0.020} & \multicolumn{1}{c|}{0.353±0.027} & 0.445±0.050 \\ \hline
4 & \multicolumn{1}{c|}{0.383±0.019} & \multicolumn{1}{c|}{0.401±0.038} & \multicolumn{1}{c|}{0.490±0.033} & \multicolumn{1}{c|}{0.380±0.025} & \multicolumn{1}{c|}{0.405±0.027} & 0.558±0.037 \\ \hline
5 & \multicolumn{1}{c|}{0.566±0.033} & \multicolumn{1}{c|}{0.597±0.030} & \multicolumn{1}{c|}{0.602±0.047} & \multicolumn{1}{c|}{0.507±0.021} & \multicolumn{1}{c|}{0.533±0.027} & 0.615±0.040 \\ \hline
6 & \multicolumn{1}{c|}{0.591±0.030} & \multicolumn{1}{c|}{0.608±0.037} & \multicolumn{1}{c|}{0.710±0.048} & \multicolumn{1}{c|}{0.521±0.020} & \multicolumn{1}{c|}{0.558±0.041} & 0.630±0.052 \\ \hline
7 & \multicolumn{1}{c|}{0.313±0.014} & \multicolumn{1}{c|}{0.388±0.041} & \multicolumn{1}{c|}{0.338±0.038} & \multicolumn{1}{c|}{0.288±0.014} & \multicolumn{1}{c|}{0.297±0.030} & 0.465±0.026 \\ \hline
8 & \multicolumn{1}{c|}{0.419±0.030} & \multicolumn{1}{c|}{0.532±0.050} & \multicolumn{1}{c|}{0.473±0.052} & \multicolumn{1}{c|}{0.418±0.020} & \multicolumn{1}{c|}{0.376±0.039} & 0.587±0.041 \\ \hline
9 & \multicolumn{1}{c|}{0.470±0.027} & \multicolumn{1}{c|}{0.458±0.051} & \multicolumn{1}{c|}{0.663±0.036} & \multicolumn{1}{c|}{0.380±0.019} & \multicolumn{1}{c|}{0.297±0.028} & 0.620±0.031 \\ \hline
10 & \multicolumn{1}{c|}{0.633±0.033} & \multicolumn{1}{c|}{0.613±0.041} & \multicolumn{1}{c|}{0.787±0.039} & \multicolumn{1}{c|}{0.426±0.029} & \multicolumn{1}{c|}{0.400±0.031} & 0.667±0.039 \\ \hline
12 & \multicolumn{1}{c|}{0.444±0.025} & \multicolumn{1}{c|}{0.505±0.037} & \multicolumn{1}{c|}{0.460±0.038} & \multicolumn{1}{c|}{0.419±0.024} & \multicolumn{1}{c|}{0.402±0.029} & 0.568±0.047 \\ \hline
 \textbf{Mean}& \multicolumn{1}{c|}{\textbf{0.503}} & \multicolumn{1}{c|}{\textbf{0.527}} & \multicolumn{1}{c|}{\textbf{0.588}} & \multicolumn{1}{c|}{\textbf{0.439}} & \multicolumn{1}{c|}{\textbf{0.430}} & \textbf{0.600} \\ \hline
\end{tabular}
\caption{Classification accuracy of phoneme articulations using model in {\em section} \ref{sec:recurrent}.  We also show classification accuracy for 23 consonant phonemes and 15 vowel phonemes separately. {\em A-}all phonemes, {\em C-}consonants only, {\em V-}vowels only. It is a 38-way classification problem with chance accuracy of $\frac{1}{38} = 0.026$. Accuracies are averaged over 10 random seeds.}
\label{tab:phonemeRNN}
\end{table*}

\begin{table*}[ht]
\centering
\footnotesize
\begin{tabular}{|c|p{3cm}|p{3cm}|}
\hline
\begin{tabular}[c]{@{}c@{}}Subject number\end{tabular} 
& \multicolumn{2}{c|}{\makecell{Decoding accuracy of word \\articulations using model in \\{\em section} \ref{sec:recurrent}}} \\ \hline
 & \multicolumn{1}{c|}{{\em Audible} words} & \multicolumn{1}{c|}{{\em Silent} words}  \\ \hline
1 & \multicolumn{1}{c|}{0.788±0.029} & \multicolumn{1}{c|}{0.669±0.055}  \\ \hline
2 & \multicolumn{1}{c|}{0.865±0.013} & \multicolumn{1}{c|}{0.745±0.025}  \\ \hline
3 & \multicolumn{1}{c|}{0.656±0.041} & \multicolumn{1}{c|}{0.542±0.033}  \\ \hline
4 & \multicolumn{1}{c|}{0.305±0.026} & \multicolumn{1}{c|}{0.381±0.034}  \\ \hline
5 & \multicolumn{1}{c|}{0.769±0.024} & \multicolumn{1}{c|}{0.712±0.037} \\ \hline
6 & \multicolumn{1}{c|}{0.767±0.034} & \multicolumn{1}{c|}{0.674±0.027}  \\ \hline
7 & \multicolumn{1}{c|}{0.260±0.006} & \multicolumn{1}{c|}{0.274±0.025}  \\ \hline
8 & \multicolumn{1}{c|}{0.755±0.038} & \multicolumn{1}{c|}{0.661±0.025}  \\ \hline
9 & \multicolumn{1}{c|}{0.690±0.025} & \multicolumn{1}{c|}{0.523±0.037}  \\ \hline
10 & \multicolumn{1}{c|}{0.846±0.026} & \multicolumn{1}{c|}{0.738±0.021}  \\ \hline
11 & \multicolumn{1}{c|}{0.781±0.022} & \multicolumn{1}{c|}{0.810±0.024}  \\ \hline
12 & \multicolumn{1}{c|}{0.681±0.035} & \multicolumn{1}{c|}{0.606±0.036}  \\ \hline
\textbf{Mean} & \multicolumn{1}{c|}{\textbf{0.694}} & \multicolumn{1}{c|}{\textbf{0.611}}  \\ \hline
\end{tabular}
\caption{Classification accuracy of word articulations using model in section \ref{sec:recurrent}. It is a 36-way classification problem with chance accuracy of $\frac{1}{36} = 0.028$. Accuracies are averaged over 10 random seeds.}
\label{tab:wordsRNN}
\end{table*}

%%%%%%%%%%%%%%%%%%%%%%%%%%%%%%%%%%%%%%%%%%%%%%%%%%%%%%%%%%%%%%%%%%%%%%%%%%%%%%%%%%%%%%%%%%%%%%%%%%%%%%%%%%%%%%%%%%%%%%%
\twocolumn
\subsection{NATO phonetic code articulations}
\label{sec:apd8}
Here, we present the subject-wise decoding accuracies of the generalizable language-spelling paradigm described in {\em section} \ref{sec:NATO}. Four subjects articulated the \textsc{rainbow} and \textsc{grandfather} passages character by character using NATO phonetic codes in a {\em silent} manner. The decoding accuracies are presented in table \ref{tab:NATOSubjectWise}.

\begin{table}[h!]
\centering
\footnotesize
\begin{tabular}{|c|cc|c|}
\hline
\begin{tabular}[c]{@{}c@{}}Subject \\ number\end{tabular} & \multicolumn{2}{c|}{\begin{tabular}[c]{@{}c@{}}Decoding accuracy using\\model in  {\em section} \ref{sec:SPDNet}\end{tabular}} & \multicolumn{1}{c|}{\begin{tabular}[c]{@{}l@{}}Decoding accuracy using\\model in {\em section} \ref{sec:recurrent}\end{tabular}} \\ \hline
1 & \multicolumn{2}{c|}{0.697} & 0.8154 \\ \hline
2 & \multicolumn{2}{c|}{0.632} & 0.7837 \\ \hline
3 & \multicolumn{2}{c|}{0.638} & 0.7337 \\ \hline
4 & \multicolumn{2}{c|}{0.556} & 0.7606 \\ \hline
\textbf{Mean} & \multicolumn{2}{c|}{\textbf {0.631}} & {\textbf {0.773}} \\ \hline
\end{tabular}
\caption{\footnotesize Top-{\em 5} decoding accuracy of spelled-out \textsc{rainbow} and \textsc{grandfather}  passage articulations using NATO phonetic codes. It is a 26-way classification problem with chance top-{\em 5} accuracy of $0.178$.}
\label{tab:NATOSubjectWise}
\end{table}

%%%%%%%%%%%%%%%%%%%%%%%%%%%%%%%%%%%%%%%%%%%%%%%%%%%%%%%%%%%%%%%%%%%%%%%%%%%%%%%%%%%%%%%%%%%%%%%%%%%%%%%%%%%%%%%%%%%%%%%
\section{EMG data distribution shift}
\label{sec:apd7}
In {\em section} \ref{sec:dataShift}, we discussed the EMG data distribution shift across individuals. We showed that instead of similar words clustering together, each subject's embeddings exhibit a stronger within-subject similarity. To demonstrate this, we created graph edge matrices from EMG signals recorded during the {\em audible} articulation of words. These edge matrices, denoted as $\mathcal{E}$, were constructed using the entire articulation duration of 1.5 seconds.

\begin{figure}[ht] 
    \centering 
    \includegraphics[width=0.4\textwidth]{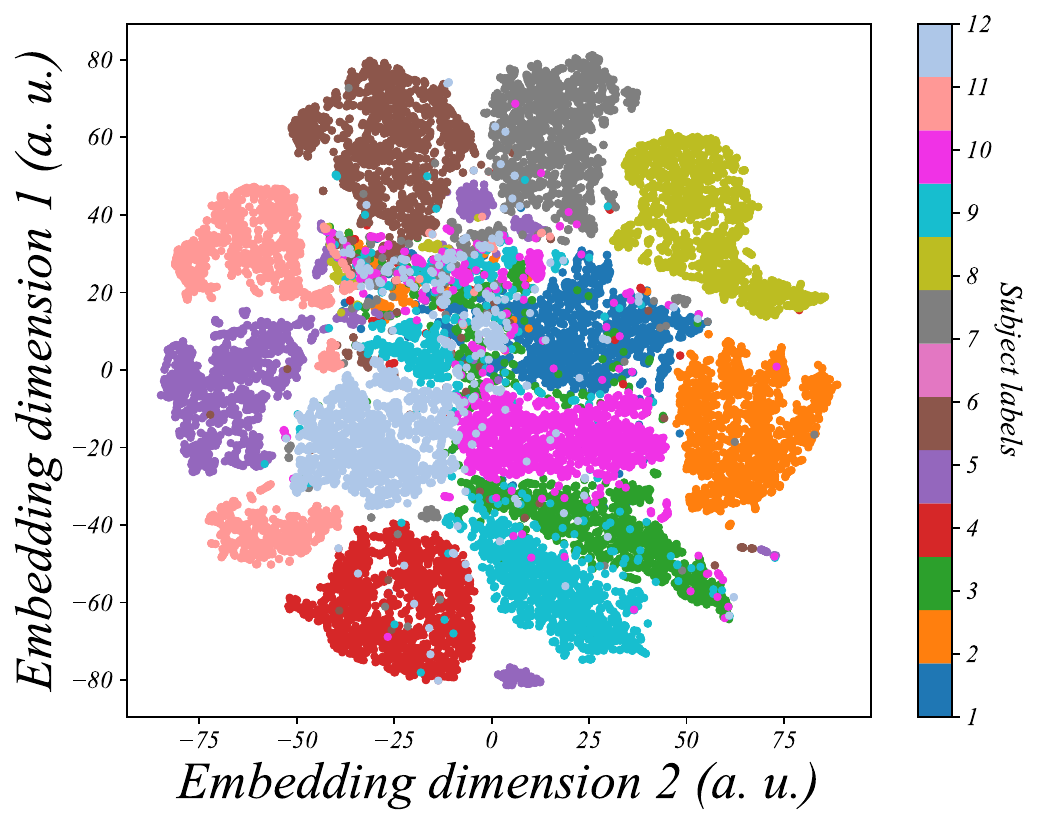} 
    \caption{Articulations from different subjects cluster separately on the manifold of SPD matrices. {\em t-}SNE of edge matrices of \textsc{rainbow} passage articulations. Embedding is colored according to subjects ({\em a.u. - }arbitrary units).} 
    \label{fig:sentencesAllIndividuals} 
\end{figure}

To further verify whether a similar distribution shift occurs on a finer temporal scale - corresponding to individual phonemes or co-articulated phonemes - we had 12 subjects articulate the \textsc{rainbow} passage {\em audibly}, as one would naturally speak. The entire passage was articulated over a duration of 215 seconds, and SPD edge matrices were constructed using a time window of 100 {\em ms}. Consequently, we obtained 2150 edge matrices for the entire passage duration, capturing speech at a phonemic level. We visualize these 2150 edge matrices from each of the 12 subjects using {\em t}-SNE ({\em section} \ref{sec:tSNE}) in figure \ref{fig:sentencesAllIndividuals}.

Similar to our observations in {\em section} \ref{sec:dataShift}, we found that articulations from different subjects exhibit strong within-subject similarity. That is, we did not observe any clustering based on articulation content - for example, similar phonemes from different subjects did not cluster together.

This finding suggests that, fundamentally, the functional connectivity of orofacial neuromuscular signals differs across individuals.

\section{Importance of electrodes in the signal graph}

In figure \ref{fig:CNN}, each column of $\mathcal{E}_{22\times 22}^{(0)}$ represents the relationship of a given node with all other nodes in the network. Since both $\mathcal{E}_{22\times 22}^{(0)}$ and $W_{22\times 22}^{(1)}$ are full-rank matrices spanning $\mathbb{R}^{22}$, each column of $W_{22\times 22}^{(1)}$ can be approximately expressed as a linear combination of the columns of $\mathcal{E}_{22\times 22}^{(0)}$. 

Formally,
\begin{equation}
    w \approx \mathcal{E}_{22\times 22}^{(0)}\kappa,
    \label{eq:importance}
\end{equation}
where $w$ denotes a column of $W_{22\times 22}^{(1)}$ and $\kappa$ is a vector of $|\mathcal{V}|$ coefficients. We estimate $\kappa$ via a least squares solution using \texttt{numpy.linalg.lstsq}. Specifically, we compute $\kappa_{\lambda_G}$ corresponding to $w_{\lambda_G}$, the eigenvector associated with the largest eigenvalue.

We repeat this procedure for all 360 articulations of \textit{audible} words (36 words from section \ref{sec:words}, each repeated 10 times). For each trial, we rank the elements of $\kappa_{\lambda_G}$ by their absolute values in descending order and identify the electrode nodes most frequently ranked among the top three across all articulations.

Our analysis reveals that the most influential nodes—i.e., those contributing most to the top eigenvector—vary substantially across subjects, indicating that the neural substrates of speech articulation are highly subject-specific. Nevertheless, certain electrode locations corresponding to the \textbf{hyoglossus}, \textbf{palatoglossus}, and \textbf{styloglossus} muscles (approximately electrodes 17, 19, 20, and 21 in figure \ref{fig:Face2}) appear frequently as important nodes across individuals. These muscles, located in the lower cheek region, play a vital role in tongue movement and are consistently recruited across a wide range of articulatory gestures.

In contrast, the importance of other orofacial muscle groups exhibits greater variability across individuals. These include muscles in the upper and posterior cheek regions—such as the \textbf{masseter} and \textbf{temporalis}, which control jaw motion, and the \textbf{zygomaticus}, involved in upper lip elevation—associated with electrode regions approximately around nodes 22, 18, and 15 in figure \ref{fig:Face2}. Electrodes located beneath the jaw capture activity from muscles involved in tongue protrusion and jaw-tongue coordination, such as the \textbf{genioglossus} (near electrodes 8 and 9 in figure \ref{fig:Face1}) and the \textbf{digastric}. Additionally, electrodes near the laryngeal region (nodes 6, 7, 10, and 11 in figure \ref{fig:Face1}) reflect the activity of muscles that modulate laryngeal and hyoid position—such as the \textbf{sternohyoid}, \textbf{stylohyoid}, and \textbf{digastric}—which are instrumental in pitch control, vowel shaping, and jaw movement.

Overall, the most informative nodes for decoding speech differ across individuals. This is consistent with the observation that the approximate eigenbasis vectors underlying articulations vary from subject to subject (see figure \ref{fig:matrixAngles}). Such variability likely stems from individual anatomical differences and personalized articulatory strategies during speech production.

In a related study \citep{gowda2025non}, where we decode {\em silently} articulated speech using EMG through phoneme-by-phoneme decoding on a limited-vocabulary corpus, we demonstrate that various subsets of electrodes perform nearly as well as the full set of electrodes.

%%%%%%%%%%%%%%%%%%%%%%%%%%%%%%%%%%%%%%%%%%%%%%%%%%%%%%%%%%%%%%%%%%%%%%%%%%%%%%%%%%%%%%%%%%%%%%%%%%%%%%%%%%%%%%%
\end{document}